\documentclass[10pt,journal,compsoc,twoside]{IEEEtran} 
\pdfoutput=1

\usepackage[nocompress]{cite}
\usepackage{./style/customstyle}

\usepackage{color}

\usepackage{xcolor}

\usepackage{url}            
\usepackage{booktabs}       
\usepackage{amsfonts}       
\usepackage{nicefrac}       
\usepackage{microtype}      
\usepackage{color}
\usepackage{caption}
\usepackage{tabulary}
\usepackage{colortbl}
\usepackage{float}
\hypersetup{
  colorlinks   = true, 
  linkcolor    = blue, 
  citecolor   = blue 
}

\graphicspath{{../pdf/}{../jpeg/}}
\DeclareGraphicsExtensions{.pdf,.jpeg,.png}

\usepackage{subfiles} 

\begin{document}

\title{Delving Into Deep Walkers: A Convergence Analysis of Random-Walk-Based Vertex Embeddings}
\author{Dominik~Kloepfer,~
        Angelica~I.~Aviles-Rivero~
        and Daniel~Heydecker
\IEEEcompsocitemizethanks{\IEEEcompsocthanksitem D. Kloepfer is with the Department
of Engineering, University of Oxford, Oxford OX1 3PJ.\protect\\
E-mail: dominik.kloepfer@eng.ox.ac.uk
\IEEEcompsocthanksitem AI Aviles-Rivero and D. Heydecker are with the Department of Applied Mathematics and Theoretical Physics, University of Cambridge, Cambridge CB3 0WA.\protect\\
E-mail: \{ai323, dh489\}@cam.ac.uk}
}

\ifCLASSOPTIONpeerreview
    \markboth{}%
    {Convergence Analysis of Random-Walk-Based Vertex Embeddings}
\else
    \markboth{}%
    {Kloepfer \MakeLowercase{\textit{et al.}}: Convergence Analysis of Random-Walk-Based Vertex Embeddings}
\fi
%

\IEEEtitleabstractindextext{%
\begin{abstract} 
Graph vertex embeddings based on random walks have become increasingly influential in recent years, showing good performance in several tasks as they efficiently transform a graph into a more computationally digestible format while preserving relevant information. However, the theoretical properties of such algorithms, in particular the influence of hyperparameters and of the graph structure on their convergence behaviour, have so far not been well-understood. In this work, we provide a theoretical analysis for random-walks based embeddings techniques. Firstly, we prove that, under some weak assumptions, vertex embeddings derived from random walks do indeed converge both in the single limit of the number of random walks $N \to \infty$ and in the double limit of both $N$ and the length of each random walk $L\to\infty$. Secondly, we derive concentration bounds quantifying the converge rate of the corpora for the single and double limits. Thirdly, we use these results to derive a heuristic for choosing the hyperparameters $N$ and $L$.
We validate and illustrate the practical importance of our findings with a range of numerical and visual experiments on several graphs drawn from real-world applications.
\end{abstract}

}

\maketitle
\IEEEpeerreviewmaketitle

\IEEEraisesectionheading{\section{Introduction}\label{sec:introduction}}
\IEEEPARstart{G}{raphs} naturally represent data arising in several real-world scenarios; important examples include social networks, medical records, protein networks and the web. In many tasks, using graph embeddings that meaningfully encode relevant information about the graph structure has enabled outstanding performance on downstream tasks, for example in node classification~\cite{neville2000iterative}, link prediction~\cite{liben2007link}, anomaly detection~\cite{akoglu2015graph}, and node clustering~\cite{nie2017unsupervised,aviles2019labelled}.

This has motivated the development of several techniques which generate embeddings by digesting the graph in different ways. One set of techniques is based on matrix factorisation, e.g.~\cite{balasubramanian2002isomap, anderson1985eigenvalues, roweis2000nonlinear, ou2016asymmetric,pang2017flexible}, in which the connections between nodes (graph properties) are represented in matrix form to obtain the embeddings. Other bodies of research have explored approaches including generative models ~\cite{le2014probabilistic,alharbi2016learning, xiao2017ssp} and hybrid (i.e. a combination of) techniques ~\cite{wei2017cross,guo2015semantically,mousavi2017hierarchical}, to name a few. 

\begin{figure}[t!]  
	\begin{centering}
		\includegraphics[width=1\linewidth]{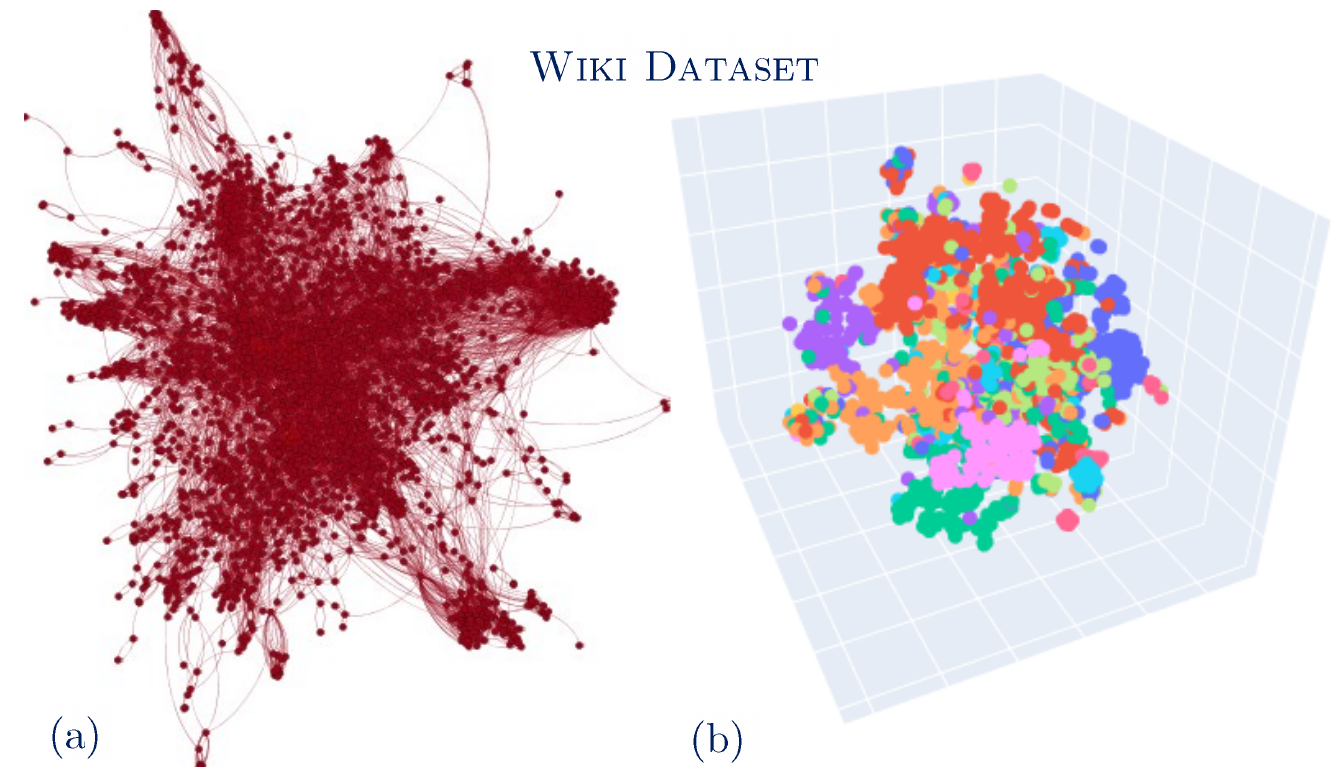}
		\par\end{centering}
	\caption{ Input-output of a random walk based vertex embeddings technique from one of our experiments. (a) Input graph from the Wiki dataset~\cite{yang2015network}. (b) Learned latent vertex representation generated by DeepWalk~\cite{perozzi2014deepwalk}. Colors indicate ground-truth classes in the dataset, and the high-dimensional embeddings have been visualised using t-SNE \cite{tsne}.} \label{fig:teaserIntroduction}
\end{figure}

Of particular interest for this paper are random-walk-based techniques, e.g.~\cite{perozzi2014deepwalk,grover2016node2vec,yang2016revisiting,cao2016dngr}, who have become popular due to their good performance. The core idea is to create a \textit{context} for each vertex by generating sequences of vertices using random walks on the graph before generating vertex embeddings from these contexts. Motivated by the success of neural language models, starting with the seminal DeepWalk~\cite{perozzi2014deepwalk} and followed by others including node2vec~\cite{grover2016node2vec} and DNGR~\cite{cao2016dngr}, a variety of algorithms view individual random walks as \say{sentences} consisting of individual \say{words} (individual vertices); a collection of random walks then corresponds to a \say{text corpus}. A corpus of random walks derived from a graph during a first stage can then be analysed in a second stage using language models such as SkipGram~\cite{mikolov2013skipgram}, which creates word (vertex) embeddings that allow one to predict a word's context from its embedding. This process is illustrated in \cref{fig:teaserIntroduction} using one of our experiments, where the graph input and the latent vertex representation are displayed.




\textbf{Contributions.} Despite their wide-spread usage, not much effort has been spent on developing a firm theoretical understanding of these algorithms. In particular, the study of the convergence of random-walk based algorithms is relatively new and incomplete.
While there are several works that slightly vary the graph sampling procedure in the first stage, e.g. \cite{grover2016node2vec} and \cite{perozzi2017don}, in this paper we mainly consider algorithms that share the sampling strategy of DeepWalk~\cite{perozzi2014deepwalk} due to this strategy's practical advantages (requiring, for example, fewer hyperparameters than node2vec~\cite{grover2016node2vec}) and because it continues to be the backbone for many new  techniques and applications, e.g.~\cite{tu2016max,qiu2018unifying,chen2019deepwalk,chen2020prediction}. 
Our contributions are:
\begin{itemize}
   \item  After formally describing DeepWalk-related techniques (\Cref{sec:formaldeepwalkdescr}), we provide a theoretical analysis for random-walk based embedding algorithms. (\Cref{sec:convergence})
   \begin{itemize}
     \item  In \cref{subsec:convrandomwalks}, we formalise the random-walk sampling strategy used by DeepWalk~\cite{perozzi2014deepwalk} in its first stage and derive the asymptotic limit of the corpora generated. We show that as the number of random walks $N$ goes to $\infty$ this convergence is almost sure both in the single limit $N \to \infty$ and the double limit $N, L \to \infty$ where $L$ is the length of each random walk. (\Cref{thm:relpairoccurrence}, \Cref{cor:longerrandomwalks})
     
     \item In \cref{subsec:convrate}, we derive concentration bounds that quantify the convergence rate of the corpora in both these cases. (\Cref{thm:Nconvspeed}, \Cref{thm:jointconvrate})
     
     \item We apply the results of this analysis to derive a heuristic that gives optimal values for the hyperparameters $N$ and $L$ for a fixed computational cost. (\Cref{cor:heuristic})
   \end{itemize}
   
   \item We validate our theoretical results 
   through a set of experiments using medium-sized and large real-world graphs consisting of thousands to millions vertices / edges. (\Cref{sec:experiments})
 \end{itemize}

\section{Related Work} \label{sec:relatedwork}

This paper is interested in random-walk-based techniques for vertex embeddings, which have demonstrated outstanding performance. We first briefly review existing such techniques.

\smallskip
\textbf{Random-Walk Based Vertex Embeddings.} The seminal work of Perozzi et al.~\cite{perozzi2014deepwalk} introduced the DeepWalk technique, which learns embeddings in two stages. In the first stage, a given graph is sampled using random walks, followed by an application of the SkipGram model to then generate vertex embeddings that allow one to predict the vertices that appear within a window around the vertex in the random walks. The promising performance of the DeepWalk motivated the developments of subsequent techniques that follow the principles of the DeepWalk.

%
Grover et al.~\cite{grover2016node2vec} introduce an additional parameter to control the random walks by allowing the user to interpolate between breadth-first and depth-first graph searches. Another modification of the first stage called Walklets was introduced in \cite{perozzi2017don}, where authors skipped over steps in each random walk. The HARP model introduced in \cite{chen2018harp} improves the initialisation of the SkipGram stage. Slightly improved performance of these models compared to DeepWalk however comes at the cost of increased computational costs.
Another set of techniques modified DeepWalk by introducing different sampling techniques for the random walks (e.g.~\cite{yang2016revisiting,pimentel2017unsupervised}), and enforcing preservation of some graph properties (e.g.~\cite{yang2015multi,pan2016tri,li2016discriminative}). Other variants of the such family of techniques are the works of~\cite{tang2015line, cao2015grarep, ou2016asymmetric, cao2016dngr}.
There exist also techniques that aim to create similar network embeddings without (explicitly) sampling any random walks, for example \cite{chen2019fastrp}.

While DeepWalk and its variants have been widely explored in the literature from a practical point of view, the analysis of the theoretical properties of such techniques is still relatively new. There are only few works that discussed the theoretical properties of this family of techniques. In the following, we discuss them and note the differences with the present work.

\smallskip
\textbf{Theoretical Analyses of DeepWalk \& Comparison to Our Work.} The first theoretical analysis of random-walk based vertex embedding techniques is due to Qiu et al.~\cite{qiu2018unifying}, who demonstrated a relationship between DeepWalk and a particular matrix factorisation problem. In the notation of \cref{sec:formaldeepwalkdescr}, that work identifies the limiting behaviour as the length of the walk $L\to\infty$, but does not prove a quantitative rate, and keeps the number of trials $N$ fixed rather than identifying the convergence as $N\to \infty$ or the convergence of the joint limit as done here. The behaviour of DeepWalk in the limit of many shorter (instead of fewer long) random walks as investigated in our work arguably has greater practical relevance as it allows for much easier parallelisation of the algorithm. Additionally, the proof in \cite{qiu2018unifying} only applies to undirected, non-bipartite graphs whereas our \Cref{thm:relpairoccurrence} and Corollaries \ref{cor:longerrandomwalks} and \ref{cor:effectstartingdistr} apply to directed and undirected graphs, both periodic and aperiodic.

More recently, another work \cite{qiu2020matrix} proved a concentration bound in terms of $L$ for arbitrary $N$ for the convergence of a corpus of random walks generated from an aperiodic graph. In contrast, we derive concentration bounds in terms of $N$ for arbitrary $L$ (\Cref{thm:Nconvspeed}) and in terms of $N$ and $L$ for the joint limit (\Cref{thm:jointconvrate}) for both periodic and aperiodic graphs. Furthermore, our bounds do not depend on the size of the graph, a factor that while of constant order can be very significant in real-world graphs with vertices and edges potentially numbering in the millions (cf. \cref{table:datasets}). In this paper, we also consider the assumptions necessary for convergence of the co-occurrence matrix to imply convergence of the extracted vertex representations (\Cref{thm:convvertexreps}), a question that \cite{qiu2020matrix} does not address.

Most recently, \cite{zhang2021consistency} considers random graphs, in either the sparse or dense regimes, in the asymptotic where the size $|V|$ and the number of sampled paths $N\to \infty$ on a fixed time interval. While also investigating the theoretical properties of random-walk based graph embedding algorithms, the setting that they consider is different to the one considered in this paper or, in fact, in~\cite{qiu2020matrix}.

Moreover and in contrast to the works above, we illustrate the practical relevance of our analysis in \Cref{cor:heuristic} by combining our results to derive a rule for the optimal choice of the hyperparameters $N$ and $L$. Qiu et al. \cite{qiu2020matrix} also derive a heuristic for choosing $L$, but they assume implicitly that $N=1$, which as our experiments in \cref{sec:experiments} show yields worse task-performance than our heuristic.
\section{Formal Description of the Deepwalk Algorithm} \label{sec:formaldeepwalkdescr}
    The DeepWalk algorithm consists of two phases: in the first phase it generates a corpus of vertex pairs, and in the second phase it executes the SkipGram algorithm (with negative sampling) on this corpus to generate vertex embeddings. 

\subsection{Generation of Corpus}

    Informally, in the first phase the corpus $\mathcal{D}$ is populated with all pairs of vertices that are within a given window size $T$ of each other in random walks generated on the graph. For each vertex, the vertices at most $T$ removed in the random walk are viewed as that vertex's \say{context} vertices, and each $(\textnormal{vertex},\; \textnormal{context vertex})$ pair is added to the corpus. 

    We take a graph $G=(V, E)$, equipped with a weight function $w:E\to [0,\infty)$.
    Consider a random walk $(V_t)_{t\ge 0}$ on $G$, with a given starting distribution $f_{V_0}$ and transition matrix $P$ with elements $P_{ij}=w_{ij}/\sum_j w_{ij}$, where $w_{ij}$ is the weight of edge $(v_i, v_j)$. We write $\textsc{Markov}(f_{V_0}, P, L)$ for the joint law of the first $L$ steps $(V_0, ...., V_{L-1})$ and, if $f_{V_0}=\delta_{v_0}$ is a one-hot distribution, write $\textsc{Markov}(v_0, P, L)$ instead. The stationary distribution of this random walk is denoted as $\pi \in \mathbb{R}^{|V|}$, which is given explicitly by $\pi(v_i)\propto\sum_j w_{ij}$.

    We define a \textit{corpus} $\mathcal{D}$ on graph $G=(V, E)$ to be a tuple $(D, m)$ consisting of the set of pairs $D = V \times V$ and the \textit{multiplicity function} $m: D \rightarrow \mathbb{N}$. Its \textit{cardinality} is the total multiplicity $|\mathcal{D}| =\sum_{d \in D} m(d)$. Since $|D|$ is finite, we may identify $m$ with the vector of its values, and write $m \in \mathbb{R}^{|V|\times|V|}_{\geq 0}$, and we denote its elements as $m(i, j) = m((v_i, v_j))$ for all $v_i, v_j \in V$, so for all $(v_i, v_j) \in D$. This matrix $m$ is also called the \textit{co-occurrence matrix}, as it encodes the number of co-occurrences of two vertices (the number of times the appear within $T$ random walk steps of each other) in the random walks performed. Note that once we have specified a graph, different corpora only differ in their multiplicity functions. 
        
    In the first phase of DeepWalk, an initially empty corpus is populated by repeatedly updating the multiplicity function. Formally, the algorithm with objective to populate the corpus $\mathcal{D}$ is given in Algorithm~\ref{algo:corpusgeneration}, taking as parameters the number of walks $N$, the walk length $L$ and window size $T$.

    In the second loop in line \ref{line:secondloop}, we follow \cite{qiu2018unifying} in having $j$ take on $L-T$ different values to avoid edge effects (when $j+r > L-1$). This simplifies the derivation in \cref{subsec:convrandomwalks} while only removing constant order terms (depending only on the fixed length $T$) from the equations.

\subsection{Generation of Vertex Representation}
    Having obtained a corpus in the first stage, in the second phase the vertex representations are calculated from the corpus. Traditionally using either the SkipGram algorithm~\cite{mikolov2013skipgram} or SkipGram with Negative Sampling~\cite{mikolov2013negativesampling} (though in principle other methods can be used as well), the vertex representations $\{z_i\}_{i:\;v_i \in V}$ are obtained by maximising an objective function of the form
    \begin{equation} \label{assumption:objfunctionform}
        F(Z, \mathcal{D}) = \sum_{(v_i, v_c)\in D} m((v_i, v_c))\cdot g_{ic}(Z)
    \end{equation}
    with respect to the $d\times|V|$ matrix $Z$. Here, $Z$ is the matrix of vertex embeddings whose columns are the individual $z_i$, $d$ is the dimension of each embedding, and $g_{ic}$ is some $g_{ic}: \mathbb{R}^{d \times|V|} \to \mathbb{R}$. 
    
    In the following, we will use $f = F(\cdot, \mathcal{D} \in \mathbb{D})$ to refer to the \textit{partial application} of an objective function where the corpus has been fixed. This partial application of $F$ yields a function $f: \mathcal{X} \to \mathbb{R}$ that calculates the objective value for a certain vertex embedding; different corpora now correspond to different $f$ for the same objective function $F$. 
    
    The second stage obtains the vertex embeddings by fixing the corpus $\mathcal{D}$ generated in the first stage from Algorithm~\ref{algo:corpusgeneration} in the objective function $F$ to create a partial application $f: \mathcal{X} \to \mathbb{R}$ which then is optimised with respect to the vertex embeddings by an optimisation procedure $A$. To keep our discussion as general as possible, we view the optimisation procedure as a deterministic function $A: f \mapsto \mathcal{X}$ which maps an objective function (with the corpus fixed) onto the space of vertex embeddings (the remaining argument of $f$); stochastic components can be incorporated by fixing the random seed of a pseudo-random number generator.

         \begin{algorithm}[t!]
           \caption{Generating Corpus $\mathcal{D}$ on graph $G=(V, E)$}
             \label{algo:corpusgeneration}
            \begin{algorithmic}[1]
                %
                %
                \STATE $m \gets \mathbf{0}_{|V|\times |V|}$
                \FOR{$n \gets 1 \textnormal{ to } N$}
                    \STATE $v_0^n \gets V_0 \sim f_{V_0}$
                    \STATE $(v_0^n, v_1^n, \dots, v_{L-1}^n) \gets \textsc{Markov}(v_0, P, L)$
                    \FOR{$j \gets 0 \textnormal{ to } L-T-1$} \label{line:secondloop}
                        \FOR{$r \gets 1 \textnormal{ to } T$}
                            \STATE $m(v_j^n, v_{j+r}^n) \gets m(v_j^n, v_{j+r}^n) + 1$
                            \STATE $m(v_{j+r}^n, v_{j}^n) \gets m(v_{j+r}^n, v_{j}^n) + 1$
                        \ENDFOR
                    \ENDFOR
                \ENDFOR
                \RETURN $\mathcal{D} = (V\times V, m)$
            \end{algorithmic}
        \end{algorithm}

\section{Theoretical Analysis}\label{sec:convergence} We can now give our main theoretical results. 
  We first prove convergence of the vertex co-occurrences as $N\to \infty$ and in the joint limit $N, L \to \infty$ in \cref{subsec:convrandomwalks}. We then quantify the rate of these convergences and show how they lead to a heuristic for choosing the hyperparameters $N, L$ in \cref{subsec:convrate}. We finally show in  \cref{subsec:convvertexreps} that, under relatively weak assumptions, convergence of the vertex co-occurrences implies the convergence of the learned vertex representation.

\subsection{Convergence of Vertex Co-Occurrences}\label{subsec:convrandomwalks}

    In this section, we derive the limit of the vertex co-occurrences (the frequencies with which a given vertex pair appears in the corpus) as the number of random walks $N$ and then also their length $L$ become infinite.

    \begin{theorem}[Occurrence Vertex Co-Occurrences]\label{thm:relpairoccurrence}
        Let $\mathcal{D}$ be a corpus generated by \Cref{algo:corpusgeneration}. Retaining the notation from that algorithm, the expected relative frequency of occurrence of the pair $(v, c)$ in the corpus is
        \begin{align*}
            \mathbb{E}\bigg[&\frac{m(v, c)}{|\mathcal{D}|}\bigg] = \\
                &\mathbb{E}_{s \sim f_{V_0}} \bigg[\frac{1}{L-T}\sum_{j=0}^{L-T-1} \frac{1}{2T}\sum_{r=1}^T (P^{j})_{sv}(P^r)_{vc} \\
                &\hspace{1cm}+ (P^{j})_{sc}(P^r)_{cv}\bigg].
        \end{align*}
        Furthermore, 
        \begin{equation*}
            \frac{m(v, c)}{|\mathcal{D}|}- \mathbb{E}\left[\frac{m(v, c)}{|\mathcal{D}|}\right] \longrightarrow 0\; \textnormal{almost surely as} \; N\rightarrow \infty.
        \end{equation*}
    \end{theorem}

    \begin{IEEEproof}
        A corpus $\mathcal{D}$ generated from $N$ random walks is the union of $N$ independent sub-corpora $\{\mathcal{D}_n\}_{1\leq n\leq N}$, which are each generated from just a single random walk. Thus the expected value of $\frac{m(v, c)}{|\mathcal{D}|}$ is the expected value of $\frac{m_n(v, c)}{|\mathcal{D}_n|}$ (the relative occurrence frequency in sub-corpus $\mathcal{D}_n$), averaged over all $n$. Since the sub-corpora are independent, we have 
        \begin{equation}
            \mathbb{E}\frac{m_n(v, c)}{|\mathcal{D}_n|} = \mathbb{E}\frac{m_m(v, c)}{|\mathcal{D}_m|} \; \forall n, m 
        \end{equation}
        and hence 
        \begin{equation}
            \mathbb{E}\frac{m(v, c)}{|\mathcal{D}|} = \mathbb{E}\frac{m_n(v, c)}{|\mathcal{D}_n|} \; \forall n.
        \end{equation}

        In the remainder of the proof we can therefore focus on a corpus $\mathcal{D}_n$ generated by just a single random walk. We again denote with $v_j$ the $j$th vertex in this random walk generating~$\mathcal{D}_n$.

        The corpus $\mathcal{D}_n$ contains all the pairs $(w, c)$ such that for $0\leq j \leq L-T-1 \;\textnormal{and}\; 1\leq r\leq T$ either $(w = v_j, c = v_{j+r})$ or $(w = v_{j+r} , c = v_j)$ (cf. \Cref{algo:corpusgeneration}). 
        
        $\mathbb{E}\frac{m_n(w, c)}{|\mathcal{D}_n|}$ is the probability with which we would draw the pair $(w, c)$ from the corpus $\mathcal{D}_n$. Since $\mathcal{D}_n$ consists of the pairs from just a single (realisation of the) random walk $(v_i)_{0\leq i\leq L-1}$, 
        \begin{equation}
        \begin{split}
            \mathbb{P}(\textnormal{draw}\;&(w, c) \;\textnormal{from}\; \mathcal{D}_n) = \\
                &\mathbb{P}(\textnormal{draw}\;(w, c) \;\textnormal{from}\; (v_i)_{0\leq i\leq L-1}).
        \end{split}
        \end{equation} 
        
        The number of pairs $(v_i, v_j)$ in the corpus with $i > j$ is the same as the number of pairs with $i < j$. For a fixed start-vertex $s$, fixed $j$, and fixed $r$, the probability of drawing the vertex pair $(w, c)$ from the random walk therefore is
        \begin{align}
        \begin{split}
            \mathbb{P}(\textnormal{draw}\;&(w, c) \; | \; s, j, r) = \\
                &\frac{1}{2}[\mathbb{P}(w = v_j , c = v_{j+r} \; | \; s, j, r) \\
                &+ \;\mathbb{P}(w = v_{j+r} , c = v_j \; | \; s, j, r)].
        \end{split}
        \end{align}

        Also,
        \begin{alignat}{2}
            \mathbb{P}(w = v_j &, c = &&v_{j+r} \; | \; s, j, r) \\
                &= \;&&\mathbb{P}(w = v_j \; | \; s = v_0) \notag \\
                &\; &&\cdot \mathbb{P}(c = v_{j+r} \; | \; w = v_j, \; s = v_0) \label{proofeqn_1}\\[2ex]
            &= \; &&\mathbb{P}(w = v_j \; | \; s = v_0) \notag \\
            & \; &&\cdot \mathbb{P}(c = v_{j+r} \; | \; w = v_j) \label{proofeqn_2}\\[2ex]
            & = &&(P^j)_{sw} \cdot (P^r)_{wc}
        \end{alignat}
        where in going from \eqref{proofeqn_1} to \eqref{proofeqn_2} we used the Markov Property for the random walk $v_i$.
        We similarly obtain  
        \begin{equation}
            \mathbb{P}(w = v_{j+r}, c = v_j \; | \; s, j, r) = (P^j)_{sc} \cdot (P^r)_{cw}\;.
        \end{equation}

        The distributions of $j$ and $r$ are uniform (the number of pairs with some $j$, $r$ in $\mathcal{D}_n$ is the same for all $i, j$), so to evaluate the expectation values with respect to $j$ and $r$ we take the arithmetic mean. From \Cref{algo:corpusgeneration} we see that the distribution of $s$ is $f_{V_0}$, so the first part of the theorem follows.

        For the second part, note that $\frac{m(v, c)}{|\mathcal{D}|}$ is exactly the mean of the independent and identially distribution random variables  $\frac{m_n(v, c)}{|\mathcal{D}_n|}$, which take values in $[0,1]$. The law of large numbers therefore applies to show the almost sure convergence claimed.
    \end{IEEEproof}

    \begin{corollary}[Effect of Longer Random Walks]\label{cor:longerrandomwalks}
        Using the ergodic theorem (cf. Theorem 1.10.2 in \cite{norris1997markovchainsbook}), 
        we can evaluate first the sum over $j$, and then the expectation value with respect to $s$, as $L\rightarrow \infty$.

        \begin{equation}\label{eqn:expectreloccurrenceconv}
            \mathbb{E}\left[\frac{m(v, c)}{|\mathcal{D}|}\right] \xrightarrow{a.s.} \; \frac{1}{2T}\sum_{r=1}^T \pi_v (P^r)_{vc} + \pi_c (P^r)_{cv}.
        \end{equation}

        Hence, as $N, L \rightarrow \infty$,
        \begin{equation}\label{eqn:reloccurrencesinfinitewalk}
            \frac{m(v, c)}{|\mathcal{D}|} \xrightarrow{a.s.} \; \frac{1}{2T}\sum_{r=1}^T \pi_v (P^r)_{vc} + \pi_c (P^r)_{cv}.
        \end{equation}
    \end{corollary} 
    \begin{remark}
        This almost sure convergence is stronger than the convergence in probability established in \cite{qiu2018unifying}. This is due to the fact that they do not let $N \rightarrow \infty$.
    \end{remark}

    \begin{corollary}[Effect of Starting Distribution]\label{cor:effectstartingdistr}
        From the definition of the stationary distribution we see that if $f_{V_0} = \pi$, 
        \begin{equation}
            \mathbb{E}_{s\sim f_{V_0}} (P^j)_{sv} = \pi_v.
        \end{equation}
        In that case, \eqref{eqn:expectreloccurrenceconv} holds as an equality for arbitrary $L>T$, and \eqref{eqn:reloccurrencesinfinitewalk} holds for arbitrary fixed $L$ as $N\to \infty.$
    \end{corollary}
    
    The results from this section, together with \cref{thm:convvertexreps}, guarantee that if the DeepWalk algorithm performs enough and long enough random walks, the learned vertex representations converge in the sense of \cref{thm:convvertexreps}.

\subsection{Convergence Rates}\label{subsec:convrate}
    
    Let us first investigate the convergence rate as $N \rightarrow \infty$ for arbitrary $L$. 

    \begin{theorem}[Convergence Rate as $N\to \infty$]\label{thm:Nconvspeed}
        The convergence of \cref{thm:relpairoccurrence} satisfies the concentration bound
        \begin{equation}
            \mathbb{P}(\left|\frac{m(v, c)}{|\mathcal{D}|}- \mathbb{E}\left[\frac{m(v, c)}{|\mathcal{D}|}\right]\right| > \varepsilon) < 2 \exp(-2 N \varepsilon^2)
        \end{equation}
        for $\varepsilon > 0$.

    \end{theorem}

    \begin{IEEEproof}
        Let again $m_n(v, c)$ be the multiplicity function of the corpus $\mathcal{D}_n$ generated from the $n$th random walk. Each random walk has the same length, so $N \cdot |\mathcal{D}_n| = |\mathcal{D}| \forall n < N$. Since $0 \leq m_n(v, c) \leq |\mathcal{D}_n|$, Hoeffding's inequality gives
        \begin{equation}
        \begin{split}
            \mathbb{P}\left(\left|\sum_{n=1}^N m_n(v, c) - \mathbb{E}[\sum_{n=1}^N m_n(v, c)]\right| > \varepsilon\right) \\
            < 2 \exp\left(-\frac{2 \varepsilon^2}{N |\mathcal{D}_n|^2}\right).
        \end{split}
        \end{equation}
        
        Dividing by $|\mathcal{D}| = N |\mathcal{D}_n|$ and using $\sum_{n=1}^N m_n(v, c) = m(v, c)$ then yields
        \begin{equation}
        \begin{split}
             \mathbb{P}\left(\left|\frac{m(v, c)}{|\mathcal{D}|}- \mathbb{E}\left[\frac{m(v, c)}{|\mathcal{D}|}\right]\right| > \frac{\varepsilon}{N |\mathcal{D}_n|}\right) \\
            < 2 \exp\left(-\frac{2 \varepsilon^2}{N |\mathcal{D}_n|^2}\right).
        \end{split}
        \end{equation}
        Re-scaling $\varepsilon \to N |\mathcal{D}_n| \varepsilon$ then yields the theorem. 
    \end{IEEEproof}


    To derive a concentration bound for the joint limit, we first need to prove two lemmas that quantify the convergence rate of $\mathbb{E}\left[\frac{m(v, c)}{|\mathcal{D}|}\right]$ as $L\to \infty$ for undirected and directed graphs. For notational simplicity, define
    \begin{equation*}
        \omega(v, c) := \frac{1}{2T}\sum_{r=1}^T \pi_v (P^r)_{vc} + \pi_c (P^r)_{cv}
    \end{equation*}
    as the value that $\mathbb{E}\left[\frac{m(v, c)}{|\mathcal{D}|}\right]$ converges to (\Cref{thm:relpairoccurrence}).


    \begin{lemma}[Convergence Rate for Undirected Graphs]\label{lemma:speedconvundirgraph}
        Given an undirected graph, let $d_i=\sum_j w_{ij}$ be the out-degrees and $D=\text{diag}(d_i)$. Let $\{\lambda_k\}$ be the ordered eigenvalues of the symmetric normalised graph Laplacian $$ L^\text{norm}=I-D^{1/2}PD^{-1/2}. $$ Then we have the convergence estimate
        \begin{equation*}
            \begin{split}
            \bigg|\mathbb{E}\bigg[&\frac{m(v, c)}{|\mathcal{D}|}\bigg] -  \omega(v, c)\bigg| \\
                &\leq \frac{1}{L-T} \frac{1-\phi_\star^{L-T}}{1-\phi_\star} \mathbb{E}_{s \sim f_{V_0}}\left[d_s^{-\frac{1}{2}}\right] \\
            &\;\;\;\;\cdot \frac{1}{2T} \sum_{r=1}^T \sqrt{d_v}(P^r)_{vc} + \sqrt{d_c}(P^r)_{cv} = O\left(\frac{1}{L}\right)
            \end{split}
        \end{equation*}
    where $\phi_\star \equiv \mu_\star = \sup\{|1-\lambda_2|, |1-\lambda_{|V|}|\}$ for non-bipartite graphs and $\phi_\star \equiv \nu_\star = \sup\{|1-\lambda_2|, |1-\lambda_{|V|-1}|\}$ for bipartite graphs and even $L-T$.
    \end{lemma}
    \begin{IEEEproof}\renewcommand{\qedsymbol}{}
        For a general undirected graph, we get by substituting for $\omega(v, c)$ and from \cref{thm:relpairoccurrence} for $ \mathbb{E}\left[\frac{m(v, c)}{|\mathcal{D}|}\right]$
        \begin{alignat}{2}
            \bigg|\mathbb{E}\bigg[&\frac{m(v, c)}{|\mathcal{D}|}\bigg] -  \omega(v, c)\bigg| \notag \\
                &= \frac{1}{2T}| \mathbb{E}_{s \sim f_{V_0}} \mathbb{E}_j &&\sum_{r=1}^T [(P^j)_{sv} - \pi_v](P^r)_{vc} \notag \\
            &  \;&&+ [(P^j)_{sc} - \pi_c](P^r)_{cv} | \\
            &\leq \frac{1}{2T}\mathbb{E}_{s \sim f_{V_0}} \sum_{r=1}^T &&\left|\mathbb{E}_j(P^j)_{sv} - \pi_v\right|(P^r)_{vc} \notag \\
            & \qquad \qquad \qquad+&&\left|\mathbb{E}_j(P^j)_{sc} - \pi_c\right|(P^r)_{cv} \label{eqn:errorandepsmixingtime}
        \end{alignat}
        by Jensen's inequality. 
        For non-bipartite graphs, we again apply Jensen's inequality and use Theorem 5.1 in \cite{lovasz1993randomwalkssurvey} to obtain
        \begin{align}
            \mathrm{LHS} &\leq \frac{1}{2T}\mathbb{E}_{s \sim f_{V_0}}\mathbb{E}_j \sum_{r=1}^T \sqrt{\frac{d_v}{d_s}}\mu_\star^j (P^r)_{vc} \notag \\
                &\qquad \qquad \qquad \qquad+ \sqrt{\frac{d_c}{d_s}}\mu_\star^j (P^r)_{cv} \\[2ex]
            &\leq \frac{1}{L-T} \frac{1-\mu_\star^{L-T}}{1-\mu_\star} \mathbb{E}_{s \sim f_{V_0}}\left[d_s^{-\frac{1}{2}}\right] \notag\\ 
                &\qquad \cdot \frac{1}{2T} \sum_{r=1}^T \sqrt{d_v}(P^r)_{vc} + \sqrt{d_c}(P^r)_{cv} 
        \end{align}

        For $L-T$ even (so that $\mathbb{E}_j$ contains an even number of terms) for bipartite graphs, we replace $\mu_\star$ with $\nu_\star$, where the difference between the two stems from the fact that $1-\lambda_{|V|}=-1$, so in the derivation of Theorem 5.1 in \cite{lovasz1993randomwalkssurvey} successive terms involving $(1-\lambda_{|V|})^t$ cancel. The Lemma then follows.

    \end{IEEEproof}

        
        

%



    \begin{lemma}[Convergence Rate for Directed Graphs]\label{lemma:speedconvdirgraph}
        For a directed graph $G$ with period $\Theta\geq 1$,
        \begin{equation}
            \left|\mathbb{E}\left[\frac{m(v, c)}{|\mathcal{D}|}\right] -  \omega(v, c)\right| = O\left(\frac{1}{L}\right).
        \end{equation}
        Furthermore, if $G$ is aperiodic ($\Theta =1$), with constants $C$ and $\alpha \in (0, 1)$ as in Theorem 4.9 in \cite{levin2017markovmixing}, 
        \begin{align}\label{eqn:aperdirgraphconverror}
            &\left|\mathbb{E}\left[\frac{m(v, c)}{|\mathcal{D}|}\right] -  \omega(v, c)\right| \notag \\
                &\qquad \leq \frac{C}{L-T} \frac{1-\alpha^{L-T}}{1-\alpha} \frac{1}{2T}\sum_{r=1}^T(P^r)_{vc} + (P^r)_{cv}.
        \end{align}
    \end{lemma}

    To prove this for both aperiodic and periodic graphs, we need the following intermediate result.
    \begin{proposition}\label{prop:periodicchainconv}
        Let $P$ be the transition matrix of a Markov chain $\{X_t\}_{t\geq 0}$ with period $\Theta$ and stationary distribution $\pi$. Let $\{\mathcal{S}_i\}_{1\leq i\leq \Theta}$ be the disjoint subsets of the state space $S$ such that $\mathbb{P}(X_{t+1}\in \mathcal{S}_{(i+1)\mod{\Theta}}|X_t \in \mathcal{S}_i)$. If vertices $s, v \in \mathcal{S}_i$ for some $i$, then there exist (explicitable) constants $\alpha \in (0, 1), \; C > 0$ such that
        \begin{equation*}
            \left|(P^{\Theta\cdot t})_{sv} - \Theta \cdot \pi_v \right| \leq C \alpha^t.
        \end{equation*}
    \end{proposition}
    \begin{IEEEproof}
       A similar observation is well-known for Markov chains in general (Theorem 4.9, \cite{levin2017markovmixing}); in the present context, it suffices to observe that $P^\Theta|_{\mathcal{S}_i}$ is an aperiodic stochastic matrix on the block $\mathcal{S}_i$, for which $\Theta \pi|_{\mathcal{S}_i}$ is an invariant probability measure, and the lemma follows from the cited result.

        
    \end{IEEEproof}
    With this proposition in hand, we are now in a position to prove the Lemma.
    \setcounter{theorem}{7}
    \begin{IEEEproof}[Proof of \Cref{lemma:speedconvdirgraph}]
        Substituting for $\omega(v, c)$ and from \Cref{thm:relpairoccurrence}  for $ \mathbb{E}\left[\frac{m(v, c)}{|\mathcal{D}|}\right]$ and using Jensen's inequality as in the proof for \Cref{lemma:speedconvundirgraph} again leads to  \eqref{eqn:errorandepsmixingtime}.

        We prove the first part of the theorem by showing that for a general periodic graph $\left|\mathbb{E}_j(P^j)_{sv} - \pi_v\right| = O\left(\frac{1}{L}\right)$.

        Let again $\{\mathcal{S}_i\}_{1\leq i\leq \Theta}$ be disjoint subsets of the state space $S$ as in \Cref{prop:periodicchainconv}. Without loss of generality, take $s \in \mathcal{S}_i$ and $v \in \mathcal{S}_k$ with $k \geq i$, and let $a = k -i$ so that $\sum_{s' \in \mathcal{S}_k} (P^a)_{ss'} = 1$. We decompose the path from $s$ to $v$ in $j$ steps, $(P^j)_{sv}$, into a part from $s\in \mathcal{S}_i$ to an $s' \in \mathcal{S}_k$ and a part from $s'\in \mathcal{S}_k$ to $v \in \mathcal{S}_k$: 
        \begin{align}
            \mathbb{E}_j(P^j)_{sv} &= \frac{1}{L-T}\sum_{j=0}^{L-T-1}(P^j)_{sv}\\
            &= \frac{1}{L-T}\sum_{j=a}^{L-T-1} \sum_{s' \in \mathcal{S}_k} (P^a)_{ss'} (P^{j-a})_{s'v} \\
            &= \frac{1}{L-T}\sum_{n=0}^{\lfloor \frac{L-T-a-1}{\Theta}\rfloor } (P^{\Theta \cdot n})_{s'v}
        \end{align}
        since if $j < a$ then $(P^j)_{sv} = 0$, and if $(j-a)\bmod \Theta \neq 0$ then $(P^{j-a})_{s'v} =0$, and since $\sum_{s' \in \mathcal{S}_k} (P^a)_{ss'} = 1$. We can use \Cref{prop:periodicchainconv} to obtain upper and lower bounds for $(P^{\Theta \cdot n})_{s'v}$
        \begin{equation}
            \Theta\cdot\pi_v - C \alpha^n \leq (P^{\Theta \cdot n})_{s'v} \leq \Theta\cdot\pi_v + C \alpha^n
        \end{equation}
        for some constants $C$ and $\alpha \in (0, 1)$.
        %
        If $\mathbb{E}_j(P^j)_{sv} - \pi_v \geq 0$, we use the upper bound on $(P^{\Theta \cdot n})_{s'v}$, and if $\mathbb{E}_j(P^j)_{sv} - \pi_v \leq 0$ we use the lower bound to obtain:
        \begin{align}
            \mathrm{LHS} \leq &\left| \frac{1}{L-T}\sum_{n=0}^{\lfloor \frac{L-T-a-1}{\Theta}\rfloor} (\Theta\cdot\pi_v)- \pi_v\right| \notag \\
            &+ \frac{1}{L-T}\sum_{n=0}^{\lfloor \frac{L-T-a-1}{\Theta}\rfloor} (C\alpha^n).
        \end{align}
        Using the fact that $\left\lfloor \frac{x}{y} \right\rfloor = \frac{x}{y} - \frac{x\bmod y}{y}$ to evaluate the prefactor of $\pi_v$, we continue
        \begin{alignat}{2}
            \mathrm{LHS} &\leq &&\left|\frac{\Theta}{L-T} \left(\left\lfloor \frac{L-T-a-1}{\Theta}\right\rfloor + 1\right) -1 \right| \pi_v \notag \\
            & &&+ \frac{1}{L-T}\sum_{n=0}^{\lfloor \frac{L-T-a-1}{\Theta}\rfloor} C \alpha^n \\
            &\leq &&\left|\frac{\Theta - a - 1 - (L-T-a-1)\bmod \Theta}{L-T} \right| \cdot\pi_v \notag \\
            & && + \frac{1}{L-T}\sum_{n=0}^{\lfloor \frac{L-T-a-1}{\Theta}\rfloor} C \alpha^n \label{eqn:periodicgraphbound} \\
            &= && \;O\left(\frac{1}{L}\right)
        \end{alignat}
        since the remaining geometric sum is bounded by a constant.
        
        To prove the second part of the theorem, note that for aperiodic graphs, $\Theta = 1$ and $i=k \Rightarrow a = 0$. For aperiodic graphs, $C$ and $\alpha$ still have the same meaning as before. 

        Therefore, in \eqref{eqn:periodicgraphbound} the prefactor of $\pi_v$ vanishes for aperiodic graphs. The upper bound on $\left|\mathbb{E}_j(P^j)_{sv} - \pi_v \right|$  therefore evaluates to
        \begin{align}
            \left|\mathbb{E}_j(P^j)_{sv} - \pi_v \right| &\leq \frac{C}{L-T} \sum_{n=0}^{L-T-1}\alpha^n \\
            &\leq \frac{C}{L-T} \frac{1-\alpha^{L-T}}{1-\alpha}.
        \end{align}
        Substituting into \eqref{eqn:errorandepsmixingtime}, the theorem follows.
    \end{IEEEproof}
    \setcounter{theorem}{5}
    
    Finally we state and prove our theorem for the convergence rate in the joint limit $N, L \to \infty$:
    \begin{theorem}[Convergence Rate as $N, L\to \infty$]\label{thm:jointconvrate}
        The convergence of \cref{cor:longerrandomwalks} satisfies the concentration bound
        \begin{equation}
            \mathbb{P}\left(\left|\frac{m(v, c)}{|\mathcal{D}|} - \omega(v, c)\right| > \varepsilon\right) < 2 \exp(-2N(\varepsilon - U)^2)
        \end{equation}
        for $\varepsilon>U >0$, and where $U$ is the appropriate upper bound for $\bigg|\mathbb{E}\bigg[\frac{m(v, c)}{|\mathcal{D}|}\bigg] -  \omega(v, c)\bigg|$ as derived in lemmas \ref{lemma:speedconvundirgraph} and \ref{lemma:speedconvdirgraph} and depends only on $L, T$, and graph-specific constants.
    \end{theorem}
    
    \begin{IEEEproof}
        We have by the triangle inequality 
        
        \begin{align}
        \begin{split}
            &\left|\frac{m(v, c)}{|\mathcal{D}|}- \mathbb{E}\left[\frac{m(v, c)}{|\mathcal{D}|}\right]\right| \notag \\
                &\;\;\;\;\geq \left|\frac{m(v, c)}{|\mathcal{D}|}- \omega(v, c)\right| - \left| \mathbb{E}\left[\frac{m(v, c)}{|\mathcal{D}|}\right] - \omega(v, c)\right|.
        \end{split}
        \end{align} 
        
        Therefore, 
        
        \begin{align}
        \begin{split}
            \mathbb{P}&\left(\left|\frac{m(v, c)}{|\mathcal{D}|}- \mathbb{E}\left[\frac{m(v, c)}{|\mathcal{D}|}\right]\right| > \varepsilon \right) \notag \\
                &\geq \mathbb{P}\left(\left|\frac{m(v, c)}{|\mathcal{D}|}- \omega(v, c)\right| - \left| \mathbb{E}\left[\frac{m(v, c)}{|\mathcal{D}|}\right] - \omega(v, c)\right| > \varepsilon \right) \\
                & = \mathbb{P}\left(\left|\frac{m(v, c)}{|\mathcal{D}|}- \omega(v, c)\right| > \varepsilon + \left|\mathbb{E}\left[\frac{m(v, c)}{|\mathcal{D}|}\right] - \omega(v, c)\right|\right) \\
                & \geq \mathbb{P}\left(\left|\frac{m(v, c)}{|\mathcal{D}|}- \omega(v, c)\right| > \varepsilon + U\right).
        \end{split}
        \end{align}
        
        Substituting $\varepsilon \to \varepsilon - U$ and using \Cref{thm:Nconvspeed} then yields the result.
    \end{IEEEproof}

    We can use \Cref{thm:jointconvrate} to obtain a rough heuristic for an "optimal" choice of the parameters $N$ and $L$ given a fixed computational cost; the window size $T$ determines the size of the local structures which the vertex embeddings encode and is therefore usually fixed for a given application. The total computational cost of computing the co-occurrence matrix scales linearly with the number of random-walk steps $N \cdot L$, but since $L\geq T$ we define the \textit{excess computational cost} $K = N \cdot L - N \cdot T = N \cdot (L-T)$. 
    
    For notational simplicity we also denote the upper bound for the failure probability in \Cref{thm:jointconvrate} as $\delta = 2 \exp(-2N(\varepsilon - U)^2)$ and absorb all terms in $U$ that are independent of $N, L$ into a constant $g$ for notational simplicity. Then we have:
    
    \begin{corollary}\label{cor:heuristic}
        Assuming that $K \gg N$ so that $\phi^{K/N} \approx 0$ and $\alpha^{K/N} \approx 0$ in $U$, the value for $N$ that minimises the error $\varepsilon = U + \sqrt{\frac{1}{2N}\log{\frac{2}{\delta}}}$ given $K, \delta$ fixed is
        \begin{equation}
            N \approx \frac{1}{2}\sqrt[3]{\frac{K^2}{g^2}\log{\frac{2}{\delta}}}.
        \end{equation}
    \end{corollary}
    This heuristic minimises the error $\varepsilon$ given an excess computational cost $K$ and an upper bound for the failure probability $\delta$, which both can be freely chosen.
    We see that under the heuristic, $N = O\left(K^{\frac{2}{3}}\right)$, which justifies the assumption $K \gg N$. Even without this assumption, we still have an upper bound, as the neglected terms $-\phi^{K/N}, -\alpha^{K/N}$ can only make $\varepsilon$ smaller. 
    In practice, one could use $g \approx 1$ as a first approximation to the remaining constant $g$, which depends on the graph structure through the spectrum or through $\alpha$. If a more precise estimate for $g$ is needed, let us remark that this proof (and the results recalled in the Appendix) produce a construct\emph{able} $g$, in that one can follow the steps of the proofs to explicitly find $g$ for a given example. In the case where we use \Cref{lemma:speedconvundirgraph}, let us remark that many techniques are known (\cite{diaconis1991geometric}, and references therein) for bounding spectral gaps, which would lead to a constructable upper bound for $g$.

\subsection{Convergence of Vertex Representations}\label{subsec:convvertexreps}
    We finally show that convergence of the corpora, suitably normalised, is indeed sufficient to show convergence of the vertex representations.


    We first remark that, if two corpora $\mathcal{D}^i=(D^{(i)},m^{(i)}), i=1,2$ differ only by a constant multiple so that $$ D^{(1)}=D^{(2)}; \qquad m^{(1)}=c\cdot m^{(2)} $$ for some $c>0$, then the respective partial applications of the objective function also differ only up to a constant multiple: $f^{(1)}=cf^{(2)}$, and it follows that the derived vertex embeddings \begin{equation} \label{lemma:equalrepresentationssufficientcondition} \{z^{(1)}_i\}_{i:\;v_i \in V} =\{z^{(2)}_i\}_{i:\;v_i \in V}\end{equation} also agree. To remove this redundancy, we define normalised objective functions
    \begin{equation}
        f^{\textrm{norm}} = \frac{1}{|\mathcal{D}|}\cdot f. \label{eqn:normalisedobjfunctiondefn}
    \end{equation}
    By \eqref{lemma:equalrepresentationssufficientcondition} this normalisation does not change the learned representations.
    
    We next assume the following:
    \begin{assumption}\label{assumption:objfunction}
        The objective function is of the same form as in~\eqref{assumption:objfunctionform}, with the $g_{ic}$ bounded.
    \end{assumption}
    \begin{assumption}\label{assumption:optimcontinuity} 
        The map 
        \begin{equation}        
            f\mapsto f(A(f)) \text{ is continuous} 
        \end{equation} on some space of partial applications of objective functions, with respect to the essential supremum distance $\|\cdot\|_\infty$, where $A$ is the optimisation procedure $A: f \mapsto \mathcal{X}$ as before.
    \end{assumption}
    Informally, \Cref{assumption:optimcontinuity} means that if the partial application of the objective function, $f$, changes slightly, the optimisation procedure will find vertex embeddings that have a similar objective value.

    \begin{theorem}[Convergence of Vertex Representations]\label{thm:convvertexreps}
        Let $(\mathcal{D}_n)$ be a sequence of corpora generated by \Cref{algo:corpusgeneration} from the same graph such that the respective vertex pair occurrence frequencies $\left(\frac{m_n}{|\mathcal{D}_n|}\right)$ converge to the vertex pair occurrence frequencies $\frac{m}{|\mathcal{D}|}$ in some corpus $\mathcal{D}$ generated from that graph. Let $f_n$ be the partial application of an objective function given corpus $\mathcal{D}_n$, and $f^{\textnormal{norm}}_n$ the normalised objective function, and similarly define $f$ and $f^{\textnormal{norm}}$; . Then 
        \begin{equation*}
            f^{\textnormal{norm}}_n(A(f_n)) \longrightarrow f^{\textnormal{norm}}(A(f)).
        \end{equation*}
    \end{theorem}

    \begin{IEEEproof}
        The map $\frac{m(\cdot)}{\mathcal{D}}\to f^\textrm{norm}$ is continuous with respect to the uniform distance on functions, as the $g_{ic}$ are bounded.
        
        By continuity of $f^{\textrm{norm}}_n$ with respect to $\left(\frac{m_n}{|\mathcal{D}_n|}\right)$, $\left(\frac{m_n}{|\mathcal{D}_n|}\right) \rightarrow \frac{m}{|\mathcal{D}|}$ implies $f^{\textrm{norm}}_n \rightarrow f^{\textrm{norm}}$.

        By continuity of $f^{\textrm{norm}}_n(A(f^{\textrm{norm}}_n))$ with respect to $f^{\textrm{norm}}_n$, $f^{\textrm{norm}}_n \rightarrow f^{\textrm{norm}}$ implies  $f^{\textrm{norm}}_n(A(f^{\textrm{norm}}_n)) \rightarrow f^{\textrm{norm}}(A(f^{\textrm{norm}}))$.

        From~\eqref{lemma:equalrepresentationssufficientcondition} we know that $A(f^{\textrm{norm}}_n) = A(f_n)$ and $A(f^{\textrm{norm}}) = A(f)$ and the theorem follows.
    \end{IEEEproof}

    \Cref{thm:convvertexreps} means that if the vertex pair occurrence frequencies of $(\textnormal{vertex}, \textnormal{context})$ pairs converge to those of some corpus $\mathcal{D}$, then the vertex representations converge to values that have the same (normalised) objective function value as the representations derived from corpus $\mathcal{D}$ and should be similarly useful for downstream tasks.

\section{Experimental Results}\label{sec:experiments}
In this section, we detail the set of experiments that we conducted to illustrate our theoretical findings. 

\begin{table}[t!]
\caption{Characteristics of the datasets used in our experiments.}
\resizebox{0.48\textwidth}{!}{
\begin{tabular}{cccc}
\hline
\rowcolor[HTML]{EFEFEF} 
\textsc{Dataset} & \begin{tabular}[c]{@{}c@{}}\#  \textsc{of }\\ Classes\end{tabular}  & \begin{tabular}[c]{@{}c@{}}\# \textsc{of Nodes}\\ $|V|$\end{tabular} & \begin{tabular}[c]{@{}c@{}}\# \textsc{of Edges}\\ $|E|$\end{tabular} \\ \hline
\textsc{Cora} & 7 & 2,708 & 5,429 \\
\textsc{BlogCatalog} & 39 & 10,312 & 333,983 \\
\textsc{Wiki} & 19 & 2,405 & 17,981 \\
\textsc{Facebook Large} & 4 & 22,470 & 171,002 \\
\textsc{Youtube} & 47 & 1.1M & 4.9M \\ \hline
\end{tabular}
}
\label{table:datasets}
\end{table}

\begin{figure}[t!]  
		\includegraphics[width=1\linewidth]{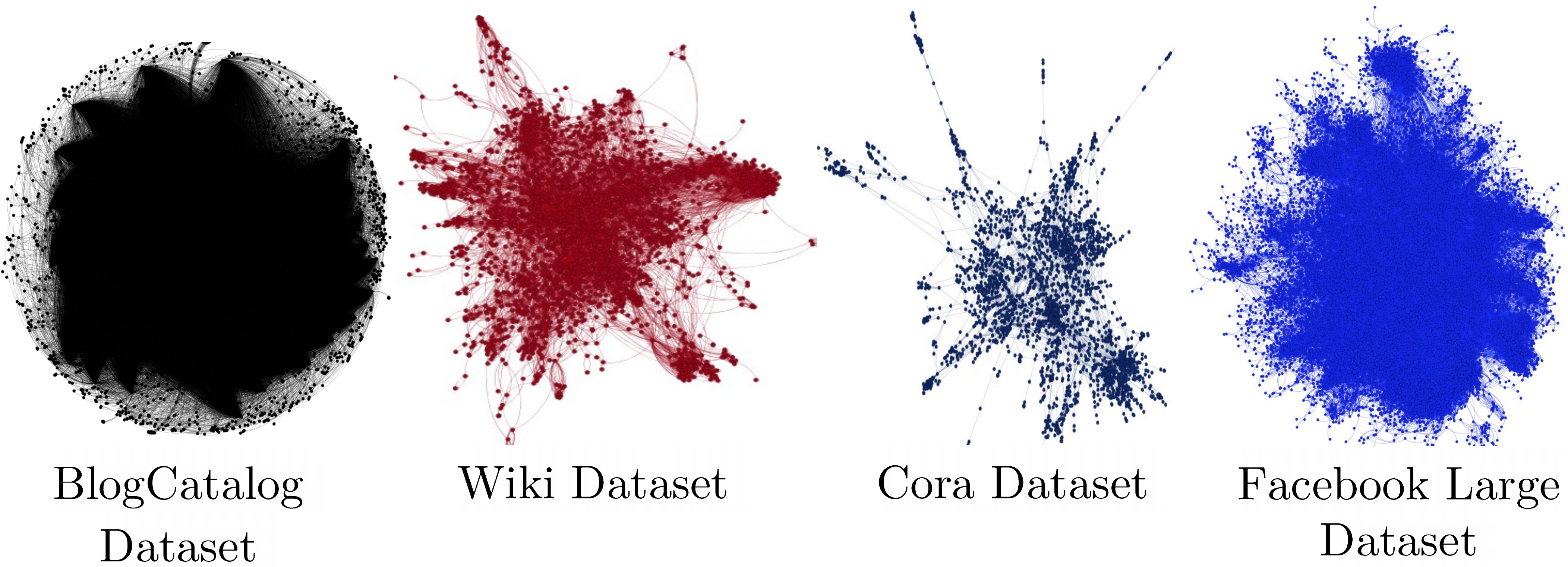}\vspace{-0.3cm}
		\vspace{-0.2cm}
	\caption{Graph visual representation of some of the datasets used in our experimental results. } \label{fig:Graphs}
\end{figure}

\begin{figure*}[t!]
\begin{minipage}[b]{0.49\linewidth}
            \centering
            \includegraphics[width=\textwidth]{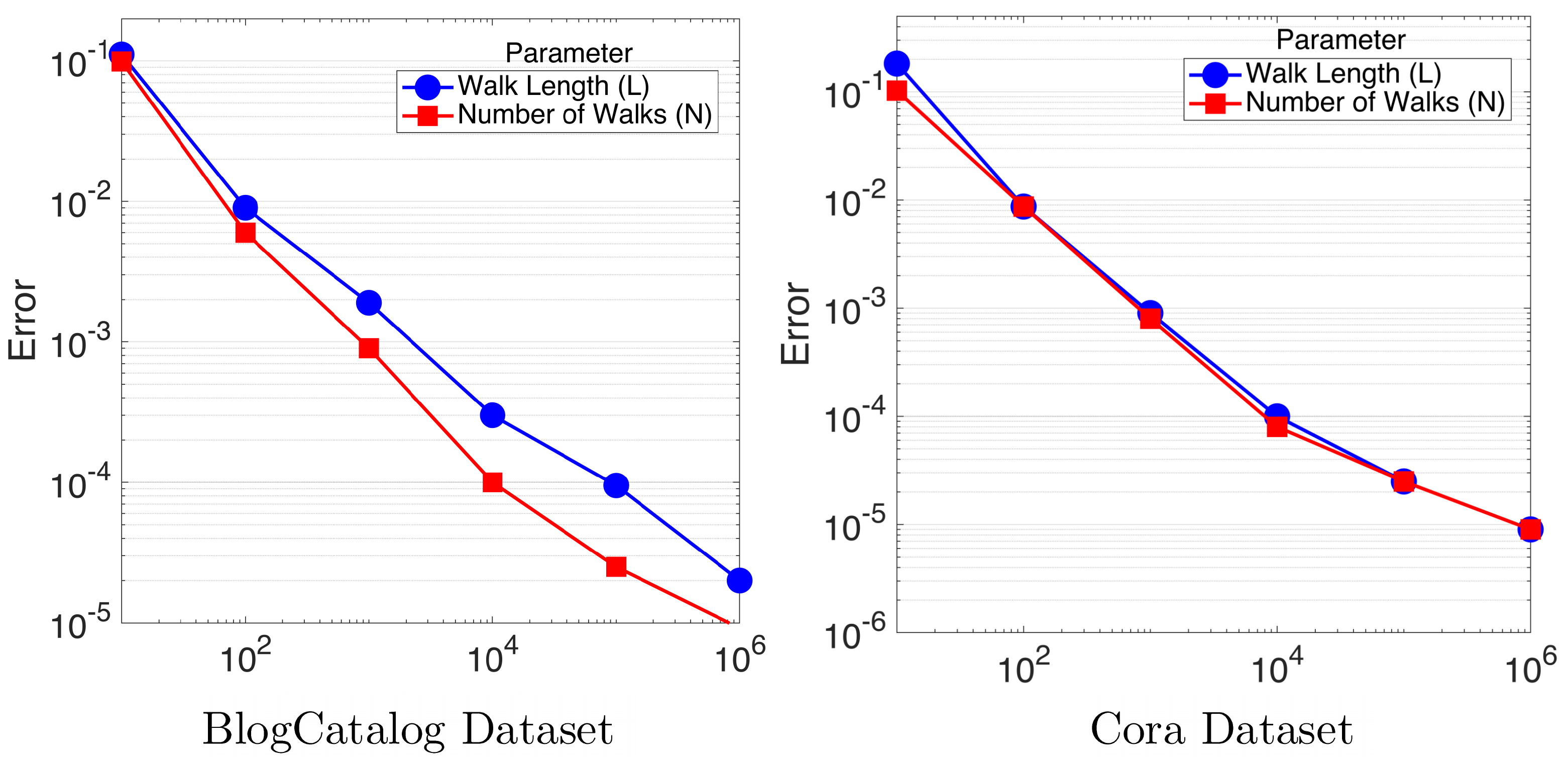}
            \caption{Convergence of co-occurrence matrices as the trajectory length $L$ and the number of walks $N$ approach infinity. The error is the computed as $\Vert m - \omega\Vert_2$, and we use $N=80$ when varying $L$ and $L=40$ when varying $N$. }
            \label{fig:matrixconvergence}
        \end{minipage}
        \hspace{0.1cm}
        \begin{minipage}[b]{0.49\linewidth}
            \centering
            \includegraphics[width=\textwidth]{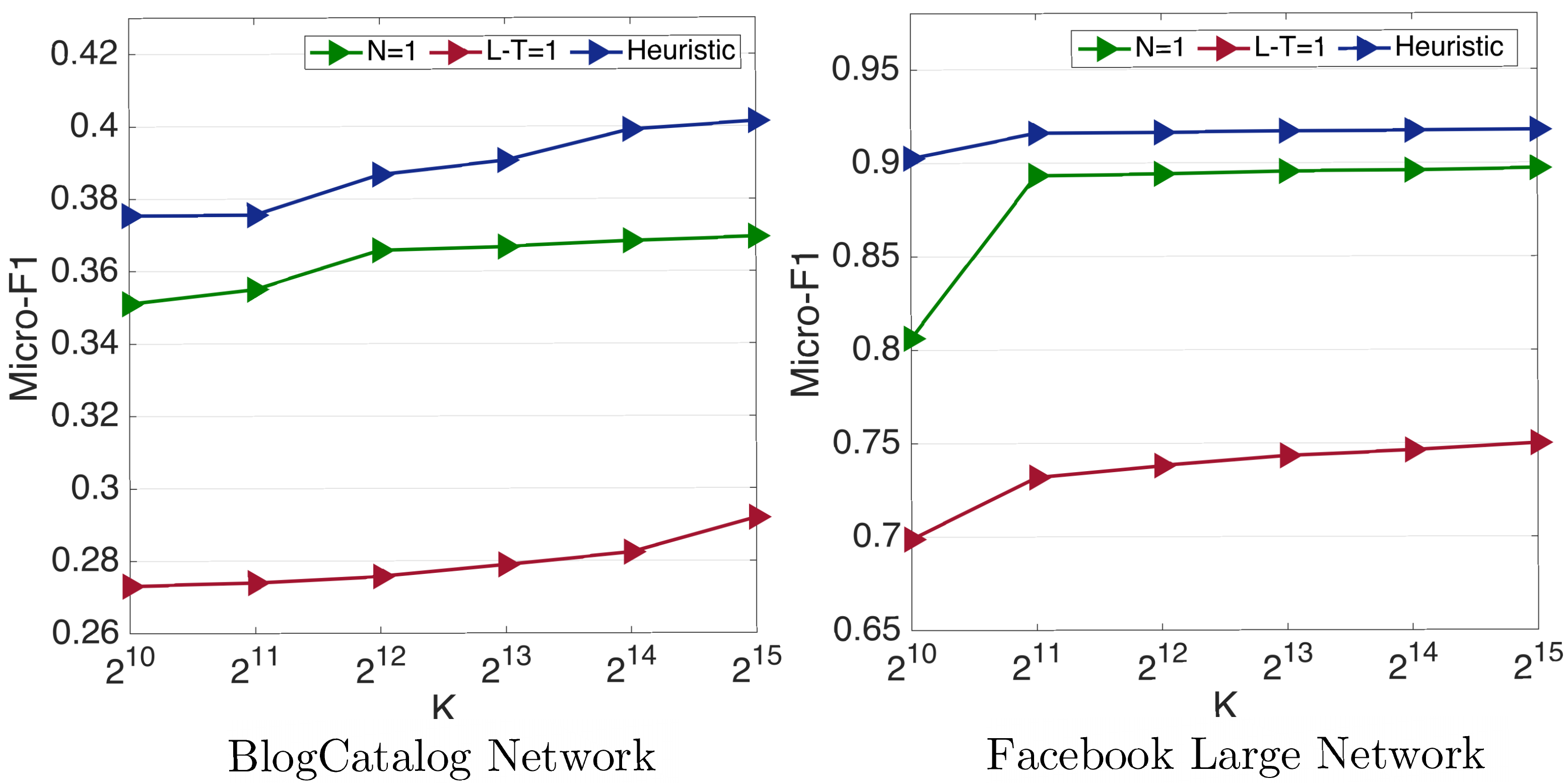}
            \caption{Computational cost in the x-axis expressed as  $K =  N \cdot (L-T)$ against the task performance, in terms of Micro-F1, on the y-axis. The plots compare our derived heuristic with two other cases where the parameters are not optimal.}
            \label{fig:heuristic}
\end{minipage}
\end{figure*}

\begin{figure*}[t!]  
	\begin{centering}
		\includegraphics[width=1\linewidth]{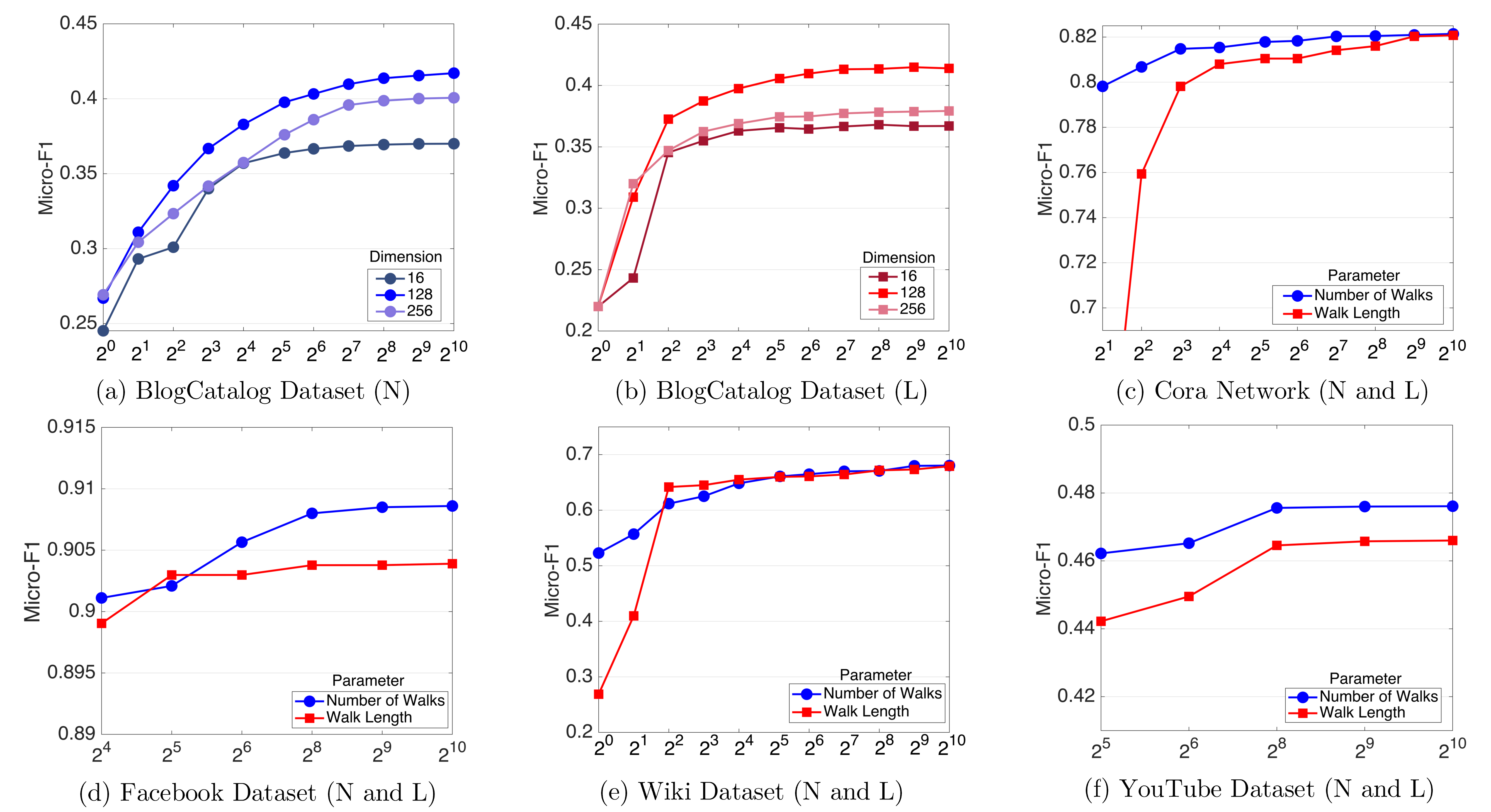}
		\par\end{centering}
	\caption{Performance on a classification task for different using embeddings generated with different hyperparameter choices for the number of random walks $N$, the length of each random walk $L$, and the embedding dimension $d$. We use $N=80$ when varying $L$ and $L=40$ when varying $N$, and fix $d=128$ for plots (c)-(f).} \label{fig:taskperformanceconvergence}
\end{figure*}

\subsection{Datasets Description}~\label{exp:eval}
We used five real-world datasets for our experiments. Each is a graph whose vertices are grouped in a number of classes. They vary in their number of nodes and edges (up to the order of millions) and in their number of classes. The dataset statistics are given in \Cref{table:datasets} along with a graphical visualisation (see \cref{fig:Graphs}).
\begin{itemize}
    \item The \textsc{BlogCatalog}~\cite{tang2009relational} dataset is derived from a blog-sharing website. The edges reflect the bloggers (vertices) following each other. The classes in the dataset are the bloggers' interests. 
    \item The \textsc{Cora}~\cite{sen2008collective} dataset is a text classification dataset for Machine Learning papers. Each vertex corresponds to a paper with the edges representing citation links. The classes are different areas in Machine Learning. The dataset also supplies bag-of-word representations for each vertex, though we did not use these in our experiments.
    \item The \textsc{Wiki}~\cite{yang2015network} dataset, similarly to Cora, is a collection of text documents (vertices) with edges representing hyperlinks between them.
    \item The \textsc{Facebook Large}~\cite{rozemberczki2019multi} dataset is a page-page graph of Facebook sites. The vertices refer to Facebook pages while the edges denote mutual likes between sites.
    \item The \textsc{YouTube}~\cite{mislove2007measurement} dataset is based on the online video-sharing platform, with edges reflecting subscription connections between users (vertices). 
\end{itemize}

\subsection{Results \& Discussion}
We support our theoretical analysis following a three-part evaluation scheme.


\smallskip
\textbf{(E1) Convergence of co-occurrence matrices.}
We begin by directly validating our results about the speed of convergence for the corpus generated by the first stage of DeepWalk (Theorems \ref{thm:Nconvspeed} and \ref{thm:jointconvrate}).

To do this, we calculate the approximation error of the co-occurrence matrix $m$ as the Frobenius norm of the difference between $m$ and its asymptotic value $\omega$, i.e. $\Vert m - \omega\Vert_2$, for different values of the random walk length $L$ and the number of random walks $N$.

More precisely, for all experiments run in this part we set the window size $T=10$ and the embedding dimension $d=128$. To observe the effect of increasing the number of walks, we set $N=\{10^1, 10^2, 10^3, 10^4, 10^5, 10^6\}$  and fix $L=40$ (see blue lines in \cref{fig:matrixconvergence}). The effect of increasing the length of the walk, we  set $L=\{10^1, 10^2, 10^3, 10^4, 10^5, 10^6\}$ and fix $N=80$ (see red lines in \cref{fig:matrixconvergence}). We ran these experiments using the \textsc{BlogCatalog} and \textsc{Cora} datasets. The results are displayed in \cref{fig:matrixconvergence}, where one can observe a convergence rate consistent with our results for the limiting cases when $N \to \infty$ and  $L \to \infty$.



\smallskip
\textbf{(E2) Heuristic for optimal values for $N$ and $L$.} We show the practical application of our heuristic for optimal choices of the hyperparameters $N$ and $L$ for a fixed computational complexity $K =  N \cdot (L-T)$ and fixed failure probability $\delta$ (\Cref{cor:heuristic}). 
For our experiments, we plot the task performance against the computational cost $K$ when choosing $N$ and $L$ using three different strategies. In the first strategy, we set $L = T + 1$ and $N = K$ to simulate the extreme of the minimum random walk length (red line in \cref{fig:heuristic}). In the second strategy, we set $N=1$ and $L = T + K$, the case of performing a single very long walk, which corresponds to the hyperparameter choice suggested in \cite{qiu2020matrix} (green line in \cref{fig:heuristic}). Finally, we select $N$ and $L$ using our heuristic, approximating the constant $g \approx 1$ and fixing $\delta = 0.01$ (blue line in \cref{fig:heuristic}).

The task performance is the Micro-F1 score achieved by the one-vs-rest logistic regression suggested by \cite{perozzi2014deepwalk} and also used in \cite{qiu2018unifying,qiu2020matrix}. For all strategies we again set $T=10$ and $d=128$. We ran these experiments using the \textsc{BlogCatalog} and the \textsc{Facebook Large} datasets for the values $K = \{2^{10}, 2^{11}, 2^{12}, 2^{13}, 2^{14}, 2^{15}\}$, so after necessary rounding the values used to test the heuristic were $N = \{88, 140, 224, 355, 563, 892\}$ and $L-T = \{12, 15, 18, 23, 29, 37\}$. The results are plotted in \cref{fig:heuristic}.

Choosing $N$ and $L$ using the heuristic consistently outperforms the other two cases, illustrating the impact that a better choice of the hyperparameters $N$ and $L$ can have. The maximally-parallelisable choice of $N=K$ and $L =T +1$ performs a lot worse than the other two strategies, which is likely due to the uniform starting distribution $f_{V_0}$ being too different to the stationary distribution on these graphs, resulting in the vertex contexts generated from the short walks not being very informative.

\textbf{(E3) Convergence of task performance.} In practice, vertex embeddings are seldom the final goal of a machine learning pipeline, and our notion of convergence in \cref{thm:convvertexreps} takes even two different sets of vertex embeddings to be equivalent if they result in the same (normalised) objective function value. We therefore use convergence of the task performance of a down-stream algorithm as a proxy for the convergence of the vertex embeddings; if the downstream task performance has converged then the vertex embeddings have very likely converged as well (in the sense of \cref{thm:convvertexreps}).

We again use the standard protocol for network embeddings evaluation (\cite{perozzi2014deepwalk,qiu2018unifying,qiu2020matrix}) and compute the Micro-F1 score of a one-vs-rest logistic regression.   
We again set the window size $T=10$. The results are plotted in \cref{fig:taskperformanceconvergence}. For the experiments in \cref{fig:taskperformanceconvergence} (a) and (b) we use varying embedding dimensions of $d = \{16, 128, 256\}$, while we keep the embedding dimension constant at $d=128$ for the other experiments in plots (c)-(f). When varying $L$, we set $N=80$, and when varying $N$, we set $L=40$.


One observes that as the number of walks and the length of each walk are tending to large values the performance converges. 
This observation holds across all graph sizes and dimensions, for both medium-sized graphs like the \textsc{Cora} dataset and large graphs like the \textsc{BlogCatalog} dataset and embedding dimensions ranging from $d=16$ to $d=256$.



\vspace{-0.2cm}
\section{Conclusion}\label{sec:conclusion}

We presented a convergence analysis for random-walk-based vertex embeddings. Unlike existing theoretical works, we investigated the convergence of vertex co-occurrences in the generated corpus as the number of random walks $N \to \infty$ for arbitrary random walk lengths $L$ as well as the simultaneous limit $N, L\to \infty$, proving almost sure convergence and quantifying the convergence rates by deriving concentration bounds for the two limits.
We also proved that convergence of vertex co-occurrences does indeed imply convergence of the vertex embeddings themselves, a fact that to our knowledge had not previously received formal theoretical treatment.
Moreover, we provided a heuristic for choosing optimal values for the hyperparameters $N$ and $L$. We supported our theory with a set of experiments using several real-world graphs, and demonstrate the practical implication of our findings.
 
\appendix[Extended Mathematical Results]

This appendix explicitly gives the mathematical results and some proofs cited, following the notation used in the main paper. 
    
    \begin{theorem}[Ergodic Theorem; Theorem 1.10.2 in \cite{norris1997markovchainsbook}]
        Let $P$ be an irreducible transition matrix and let $\lambda$ be any distribution on the state space $V$. If $(X_t)_{t\geq 0}$ is a Markov chain with transition matrix $P$, starting distribution $\lambda$, and invariant distribution $\pi$, then
        \begin{equation}
            \frac{N_i(n)}{n}\to \frac{1}{m_i} \textnormal{ almost surely as }n\to \infty
        \end{equation}
        where $m_i = \mathbb{E}(T_i) = \frac{1}{\pi_i}$ is the expected return time to state $i \in V$ and $N_i$ is the number of visits to state $i$ before time $n$. Moreover, if $P$ is finite then for any bounded function $f: V \to \mathbb{R}$ we have
        \begin{equation*}
            \frac{1}{n}\sum_{t=0}^{n-1} f(X_t) \longrightarrow \bar{f} = \sum_{i\in S}\pi_i f_i \;\;\textnormal{almost surely as} \; n\rightarrow \infty.
        \end{equation*}
    \end{theorem}
    \begin{IEEEproof}
        If $P$ is transient, then, with probability 1, the total number of visits $N_i$ to state $i$ is finite, so
        \begin{equation}
            \frac{N_i(n)}{n} \leq \frac{N_i}{n} \to 0 = \frac{1}{m_i}.
        \end{equation}
        Suppose then that $P$ is recurrent and fix a state $i$. For $T= T_i$ the return time to state $i$ we then have $\mathbb{P}(T < \infty) = 1$ and $(X_{T+n})_{n\geq 0}$ is a Markov chain with starting distribution $\delta_i$ and independent of $X_0, X_1, \dots, X_T$ by the strong Markov property. The long-run proportion of time spent in $i$ is the same for $(X_{T+n})_{n\geq 0}$ and for $(X_{n})_{n\geq 0}$, so it suffices to consider the case $\lambda = \delta_i$.
        
        Write $S_i^{(r)}$ as the length of the $r$th excursion to $i$, i.e. $S_i^{(r)} = T_i^{(r)} - T_i^{(r-1)}$ where $T_i^{(r)}$ is the $r$th return time to $i$. The non-negative random variables $S_i^{(1)}, S_i^{(2)}, \dots$ are independent and identically distributed with $\mathbb{E}(S_i^{(r)}) = m_i$ by the strong Markov property. Now
        \begin{equation}
            S_i^{(1)} + \dots + S_i^{(N_i(n) - 1)} \leq n-1, 
        \end{equation}
        the left-hand-side being the time of the last visit to $i$ before time $n$. Also,
        \begin{equation}
            S_i^{(1)} + \dots + S_i^{(N_i(n))} \geq n, 
        \end{equation}
        the left-hand side being the time of the first visit to $i$ after time $n-1$. Hence,
        \begin{equation} \label{eqn:ergodicthmproofeqn1}
            \frac{ S_i^{(1)} + \dots + S_i^{(N_i(n) - 1)}}{N_i(n)} \leq \frac{n}{N_i(n)}
        \end{equation}
        and 
        \begin{equation} \label{eqn:ergodicthmproofeqn2}
            \frac{n}{N_i(n)} \leq \frac{S_i^{(1)} + \dots + S_i^{(N_i(n))}}{N_i(n)}.
        \end{equation}
        By the strong law of large numbers
        \begin{equation}
            \frac{S_i^{(1)} + \dots + S_i^{(n)}}{n} \to m_i \textnormal{ almost surely as } n\to \infty
        \end{equation}
        and, since $P$ is recurrent, 
        \begin{equation}
            N_i(n) \to \infty \textnormal{ almost surely as } n\to \infty.
        \end{equation}
        So, letting $n\to \infty$ in \eqref{eqn:ergodicthmproofeqn1} and \eqref{eqn:ergodicthmproofeqn2}, we get
        \begin{equation}
            \frac{n}{N_i(n)} \to m_i \textnormal{ almost surely as } n\to \infty
        \end{equation}
        which implies
        \begin{equation}
            \frac{N_i(n)}{n} \to \frac{1}{m_i} \textnormal{ almost surely as } n\to \infty.
        \end{equation}
        
        Let $f: V \to \mathbb{R}$ now be a bounded function and assume without loss of generality that $|f|\leq 1$. For any $W \subseteq V$ we have
        \begin{equation}
        \begin{split}
            &\left|\frac{1}{n}\sum_{k=0}^{n-1} f(X_k) - \bar{f} \right| = \left|\sum_{i \in V}\left(\frac{N_i(n)}{n} - \pi_i\right)f_i\right| \\
            &\leq \sum_{i \in W}\left|\frac{N_i(n)}{n} - \pi_i \right| + \sum_{i \notin W}\left|\frac{N_i(n)}{n} - \pi_i \right| \\
            &\leq \sum_{i \in W}\left|\frac{N_i(n)}{n} - \pi_i \right| + \sum_{i \notin W}\left(\frac{N_i(n)}{n} + \pi_i \right) \\
            &\leq 2 \sum_{i \in W}\left|\frac{N_i(n)}{n} - \pi_i \right| + 2 \sum_{i \notin W} \pi_i.
        \end{split}
        \end{equation}
        Given $\varepsilon > 0$, choose $W$ finite such that 
        \begin{equation}
            \sum_{i \notin W} \pi_i < \varepsilon /4.
        \end{equation}
        We proved above that $\frac{N_i(n)}{n} \to \pi_i$ almost surely, so choose an $L$ such that for $n\geq L$
        \begin{equation}
            \sum_{i \in W}\left|\frac{N_i(n)}{n} - \pi_i \right| < \varepsilon /4.
        \end{equation}
        Then, for $n\geq L$ we have 
        \begin{equation}
            \left|\frac{1}{n}\sum_{k=0}^{n-1} f(X_k) - \bar{f} \right| < \varepsilon,
        \end{equation}
        which establishes the desired convergence.
    \end{IEEEproof}
    
    \begin{proposition}[Equation 3.1 in \cite{lovasz1993randomwalkssurvey}]\label{prop:transprobseigenvalueconnection}
        Consider an undirected graph with symmetric normalised Laplacian $L^\text{norm}=I-D^{1/2}PD^{-1/2}$, transition matrix $P$, and stationary distribution $\pi$. Let the ordered eigenvalues $\{\lambda_k\}$ of $L^{\textnormal{norm}}$ have normalised eigenvectors $\{v^{(k)}\}$. Then
            \begin{equation*}\label{eqn_transprobseigenvalueconnection}
                (P^t)_{ij} = \pi_j + \sum_{k=2}^{|V|} (1-\lambda_k)^t v^{(k)}_i v^{(k)}_j \sqrt{\frac{d_j}{d_i}}.
            \end{equation*}
    \end{proposition}
    \begin{IEEEproof}
        The eigenvectors of a symmetric matrix are mutually orthogonal. We can thus write in spectral form
        \begin{equation*}
            I - L^{\textnormal{norm}} = \sum_{k=1}^{|V|}(1-\lambda_k)v^{(k)} v^{(k)T}
        \end{equation*}
        
        From the definition of $L^{\textnormal{norm}}$, and using the fact that the eigenvector $v^{(1)}$ with eigenvalue $\lambda_1 = 0$ is $v^{(1)}_i = \sqrt{\pi_i}$, we get
        
        \begin{align*}
            P^t &= D^{-\frac{1}{2}}(I - L^{\textnormal{norm}})^t D^{\frac{1}{2}} \\
                &= \sum_{k=1}^{|V|}(1 - \lambda_k)^t D^{-\frac{1}{2}}v^{(k)} v^{(k)T} D^{\frac{1}{2}} \\
                &= Q + \sum_{k=2}^{|V|}(1 - \lambda_k)^t D^{-\frac{1}{2}}v^{(k)} v^{(k)T} D^{\frac{1}{2}}.
        \end{align*}
        Using $\pi_i = d_i / \sum_{k \in V}d_k$ we have 
        \begin{equation}
        \begin{split}
            Q_{ij} &= (1-\lambda_1) (D^{-\frac{1}{2}}v^{(1)} v^{(1)T} D^{\frac{1}{2}})_{ij} \\
                    &= \frac{1}{\sqrt{d_i}}\sqrt{\pi_i}\sqrt{\pi_j}\sqrt{d_j} \\
                    &= \pi_j.
        \end{split}
        \end{equation}

        The proposition follows.
    \end{IEEEproof}
    
    \begin{theorem}[Mixing Rate of Random Walk on Undirected Graph; Theorem 5.1 in \cite{lovasz1993randomwalkssurvey}]\label{thm:boundonabsolutevalue} 
        For a random walk on an undirected graph, with $\mu_\star = \sup\{|1-\lambda_2|, |1-\lambda_{|V|}|\}$ (again, $\{\lambda_k\}$ are the ordered eigenvalues of $L^{\textnormal{norm}}$),

        \begin{equation*}
            |(P^t)_{ij} - \pi_j| \leq \sqrt{\frac{d_j}{d_i}} \mu_\star^t.
        \end{equation*}
    \end{theorem}
    \begin{IEEEproof}
        Starting from \cref{prop:transprobseigenvalueconnection}, we have
        \begin{align*}
            |(P^t)_{ij} - \pi_j| &= \left|\sum_{k=2}^{|V|} (1-\lambda_k)^t v^{(k)}_i v^{(k)}_j \sqrt{\frac{d_j}{d_i}}\right| \\
                                &\leq \sqrt{\frac{d_j}{d_i}} \mu_\star^t \cdot\left|\sum_{k=2}^{|V|}v^{(k)}_i v^{(k)}_j\right| \\
                                &\leq \sqrt{\frac{d_j}{d_i}} \mu_\star^t 
        \end{align*}
        by normalisation of the $v^{(k)}$.
    \end{IEEEproof}
    
    \begin{corollary}[Mixing Rate of Random Walk on Bipartite Graph] 
        For a random walk on a bipartite undirected graph, with $\nu_\star = \sup\{|1-\lambda_2|, |1-\lambda_{|V|-1}|\}$, denoting with $\langle \cdot \rangle_t$ an average over an even number of timesteps ($\langle f(t) \rangle_t = \frac{1}{n+1}\sum_{k = t-n}^t f(k)$ with $n$ odd), 
        \begin{equation}
            |\langle (P^t)_{ij}\rangle_t - \pi_j| \leq \sqrt{\frac{d_j}{d_i}} \langle\nu_\star^t\rangle_t.
        \end{equation}
    \end{corollary}
    \begin{IEEEproof}
        It is a fact that for a bipartite graph, $1-\lambda_{|V|} = -1$. From \cref{prop:transprobseigenvalueconnection}, we then have 
        \begin{equation*}
            \langle(P^t)_{ij}\rangle_t = \pi_j + \sum_{k=2}^{|V|-1} \langle(1-\lambda_k)^t\rangle_t \cdot v^{(k)}_i v^{(k)}_j \sqrt{\frac{d_j}{d_i}},
        \end{equation*}
        since successive terms involving $(1-\lambda_{|V|})^t = (-1)^t$ cancel when averaging over an even number of $t$.

        The rest of the proof is analogous to the proof of \cref{thm:boundonabsolutevalue}.
    \end{IEEEproof}
    
    \begin{theorem}[Convergence Theorem; Theorem 4.9 in \cite{levin2017markovmixing}]
        Suppose an irreducible and aperiodic Markov Chain with transition matrix $P\in \mathbb{R}^{|V|\times |V|}$ has stationary distribution $\pi\in \mathbb{R}^{|V|}$. Then there exist constants $\alpha \in (0, 1), \; C>0$ such that
        \begin{equation*}
            \sup_{i \in S} \|(P^t)_{i \cdot} - \pi \|_{\textnormal{TV}} \leq C \alpha^t,
        \end{equation*}
        where $\|\cdot\|_{TV}$ is the total variation distance of two distributions $\mu, \nu$ on $S$, $\|\mu - \nu\|_{TV}=\max_{A\subseteq S} |\mu(A)-\nu(A)|$.
    \end{theorem}
    \begin{IEEEproof}
        Since $P$ is irreducible and aperiodic, there exists an $r$ such that $P^r$ has strictly positive entries. Let $\Pi$ be the $|V|\times |V|$ matrix whose rows are the stationary distribution $\pi$. For sufficiently small $\delta>0$ we have
        \begin{equation}
            (P^r)_{ij} \geq \delta \pi_j
        \end{equation}
        for all $i, j = 1, 2, \dots |V|$. Let $\theta = 1-\delta$. Then the equation
        \begin{equation} \label{eqn:convthmproofeqn1}
            P^r = (1-\theta)\Pi + \theta Q
        \end{equation}
        defines a stochastic matrix $Q$.
        It is a straightforward computation to check that $M\Pi=\Pi$ for any stochastic matrix $M$ and that $\Pi M = \Pi$ for any matrix $M$ such that $\pi M = \pi$.
        
        Next, we use induction to demonstrate that 
        \begin{equation}\label{eqn:convthmproofeqn2}
            P^{rk} = (1-\theta^k)\Pi + \theta^k Q^k
        \end{equation}
        for $k\geq 1$. If $k=1$ this holds by \eqref{eqn:convthmproofeqn1}. Assuming that \eqref{eqn:convthmproofeqn2} holds for $k=n$, 
        \begin{equation}
            P^{r(n+1)} = P^{rn}P^r = [(1-\theta^n)\Pi + \theta^n Q^n]P^r.
        \end{equation}
        Distributing and expanding $P^r$ in the second term, using \eqref{eqn:convthmproofeqn1}, gives
        \begin{equation}
            P^{r(n+1)} = (1-\theta^n)\Pi P^r + (1-\theta)\theta^n Q^n \Pi + \theta^{n+1} Q^{n+1}.
        \end{equation}
        Since $\Pi P^r = \Pi$ and $Q^n \Pi = \Pi$, we have
        \begin{equation}
            P^{r(n+1)} = (1-\theta^{n+1})\Pi + \theta^{n+1} Q^{n+1}
        \end{equation}
        which establishes \eqref{eqn:convthmproofeqn2} for all $k\geq 1$.
        Multiplying by $P^j$ and rearranging terms now yields
        \begin{equation}
            P^{rk+j} -\Pi = \theta^k (Q^k P^j- \Pi).
        \end{equation}
        To complete the proof, sum the absolute values of the elements in row $i$ on both sides and divide by $2$. On the right-hand-side, the second factor is at most the largest possible total variation distance between distributions, which is 1. Hence for any $i$, we have 
        \begin{equation}
            \|(P^{rk+j})_{i \cdot} - \pi\|_{\textnormal{TV}} \leq \theta^k
        \end{equation}
        and the theorem follows.
    \end{IEEEproof}


\section*{Acknowledgments}
DK is funded by a Studentship from the UK Engineering and Physical Sciences Research Council (EPSRC). 
AIAR  gratefully  acknowledges  the  financial  support  of  the  CMIH and  CCIMI  University  of  Cambridge.  
DH is funded by EPSRC grant
EP/L016516/1 for the University of Cambridge Centre for Doctoral Training, the Cambridge Centre for Analysis.  

\bibliography{bibfile}

\begin{thebibliography}{10}
\providecommand{\url}[1]{#1}
\csname url@samestyle\endcsname
\providecommand{\newblock}{\relax}
\providecommand{\bibinfo}[2]{#2}
\providecommand{\BIBentrySTDinterwordspacing}{\spaceskip=0pt\relax}
\providecommand{\BIBentryALTinterwordstretchfactor}{4}
\providecommand{\BIBentryALTinterwordspacing}{\spaceskip=\fontdimen2\font plus
\BIBentryALTinterwordstretchfactor\fontdimen3\font minus
  \fontdimen4\font\relax}
\providecommand{\BIBforeignlanguage}[2]{{%
\expandafter\ifx\csname l@#1\endcsname\relax
\typeout{** WARNING: IEEEtran.bst: No hyphenation pattern has been}%
\typeout{** loaded for the language `#1'. Using the pattern for}%
\typeout{** the default language instead.}%
\else
\language=\csname l@#1\endcsname
\fi
#2}}
\providecommand{\BIBdecl}{\relax}
\BIBdecl

\bibitem{neville2000iterative}
J.~Neville and D.~Jensen, ``Iterative classification in relational data,''
  \emph{Proc. AAAI-2000 workshop on learning statistical models from relational
  data}, 2000.

\bibitem{liben2007link}
D.~Liben-Nowell and J.~Kleinberg, ``The link-prediction problem for social
  networks,'' \emph{Journal of the American society for information science and
  technology}, vol.~58, no.~7, pp. 1019--1031, 2007.

\bibitem{akoglu2015graph}
L.~Akoglu, H.~Tong, and D.~Koutra, ``Graph based anomaly detection and
  description: a survey,'' \emph{Data mining and knowledge discovery}, vol.~29,
  no.~3, pp. 626--688, 2015.

\bibitem{nie2017unsupervised}
F.~Nie, W.~Zhu, and X.~Li, ``Unsupervised large graph embedding,'' in
  \emph{Proceedings of the AAAI Conference on Artificial Intelligence},
  vol.~31, no.~1, 2017.

\bibitem{aviles2019labelled}
A.~I. Aviles-Rivero, N.~Papadakis, R.~Li, S.~M. Alsaleh, R.~T. Tan, and C.-B.
  Schonlieb, ``When labelled data hurts: Deep semi-supervised classification
  with the graph 1-laplacian,'' \emph{arXiv preprint arXiv:1906.08635}, 2019.

\bibitem{balasubramanian2002isomap}
M.~Balasubramanian, E.~L. Schwartz, J.~B. Tenenbaum, V.~de~Silva, and J.~C.
  Langford, ``The isomap algorithm and topological stability,'' \emph{Science},
  vol. 295, no. 5552, 2002.

\bibitem{anderson1985eigenvalues}
W.~N. Anderson~Jr and T.~D. Morley, ``Eigenvalues of the laplacian of a
  graph,'' \emph{Linear and multilinear algebra}, vol.~18, no.~2, pp. 141--145,
  1985.

\bibitem{roweis2000nonlinear}
S.~T. Roweis and L.~K. Saul, ``Nonlinear dimensionality reduction by locally
  linear embedding,'' \emph{science}, vol. 290, no. 5500, pp. 2323--2326, 2000.

\bibitem{ou2016asymmetric}
M.~Ou, P.~Cui, J.~Pei, Z.~Zhang, and W.~Zhu, ``Asymmetric transitivity
  preserving graph embedding,'' in \emph{Proceedings of the 22nd ACM SIGKDD
  international conference on Knowledge discovery and data mining}, 2016, pp.
  1105--1114.

\bibitem{pang2017flexible}
T.~Pang, F.~Nie, and J.~Han, ``Flexible orthogonal neighborhood preserving
  embedding.'' in \emph{IJCAI}, 2017, pp. 2592--2598.

\bibitem{le2014probabilistic}
T.~M. Le and H.~W. Lauw, ``Probabilistic latent document network embedding,''
  in \emph{2014 IEEE International Conference on Data Mining}, 2014, pp.
  270--279.

\bibitem{alharbi2016learning}
B.~Alharbi and X.~Zhang, ``Learning from your network of friends: A trajectory
  representation learning model based on online social ties,'' in \emph{2016
  IEEE 16th International Conference on Data Mining (ICDM)}, 2016, pp.
  781--786.

\bibitem{xiao2017ssp}
H.~Xiao, M.~Huang, L.~Meng, and X.~Zhu, ``Ssp: semantic space projection for
  knowledge graph embedding with text descriptions,'' in \emph{Proceedings of
  the AAAI Conference on Artificial Intelligence}, vol.~31, no.~1, 2017.

\bibitem{wei2017cross}
X.~Wei, L.~Xu, B.~Cao, and P.~S. Yu, ``Cross view link prediction by learning
  noise-resilient representation consensus,'' in \emph{Proceedings of the 26th
  International Conference on World Wide Web}, 2017, pp. 1611--1619.

\bibitem{guo2015semantically}
S.~Guo, Q.~Wang, B.~Wang, L.~Wang, and L.~Guo, ``Semantically smooth knowledge
  graph embedding,'' in \emph{Proceedings of the 53rd Annual Meeting of the
  Association for Computational Linguistics and the 7th International Joint
  Conference on Natural Language Processing (Volume 1: Long Papers)}, 2015, pp.
  84--94.

\bibitem{mousavi2017hierarchical}
S.~F. Mousavi, M.~Safayani, A.~Mirzaei, and H.~Bahonar, ``Hierarchical graph
  embedding in vector space by graph pyramid,'' \emph{Pattern Recognition},
  vol.~61, pp. 245--254, 2017.

\bibitem{yang2015network}
C.~Yang, Z.~Liu, D.~Zhao, M.~Sun, and E.~Y. Chang, ``Network representation
  learning with rich text information.'' in \emph{IJCAI}, vol. 2015, 2015, pp.
  2111--2117.

\bibitem{perozzi2014deepwalk}
B.~Perozzi, R.~Al-Rfou, and S.~Skiena, ``Deepwalk: Online learning of social
  representations,'' in \emph{Proceedings of the 20th ACM SIGKDD international
  conference on Knowledge discovery and data mining}, 2014, pp. 701--710.

\bibitem{tsne}
L.~Van~der Maaten and G.~Hinton, ``Visualizing data using t-sne.''
  \emph{Journal of machine learning research}, vol.~9, no.~11, 2008.

\bibitem{grover2016node2vec}
A.~Grover and J.~Leskovec, ``node2vec: Scalable feature learning for
  networks,'' in \emph{Proceedings of the 22nd ACM SIGKDD international
  conference on Knowledge discovery and data mining}, 2016, pp. 855--864.

\bibitem{yang2016revisiting}
Z.~Yang, W.~Cohen, and R.~Salakhudinov, ``Revisiting semi-supervised learning
  with graph embeddings,'' in \emph{International conference on machine
  learning}.\hskip 1em plus 0.5em minus 0.4em\relax PMLR, 2016, pp. 40--48.

\bibitem{cao2016dngr}
S.~Cao, W.~Lu, and Q.~Xu, ``Deep neural networks for learning graph
  representations,'' in \emph{Proceedings of the AAAI Conference on Artificial
  Intelligence}, vol.~30, no.~1, 2016.

\bibitem{mikolov2013skipgram}
T.~Mikolov, K.~Chen, G.~S. Corrado, and J.~Dean, ``Efficient estimation of word
  representations in vector space,'' \emph{CoRR}, vol. abs/1301.3781, 2013.

\bibitem{perozzi2017don}
B.~Perozzi, V.~Kulkarni, H.~Chen, and S.~Skiena, ``Don't walk, skip! online
  learning of multi-scale network embeddings,'' in \emph{Proceedings of the
  2017 IEEE/ACM International Conference on Advances in Social Networks
  Analysis and Mining 2017}, 2017, pp. 258--265.

\bibitem{tu2016max}
C.~Tu, W.~Zhang, Z.~Liu, M.~Sun \emph{et~al.}, ``Max-margin deepwalk:
  Discriminative learning of network representation.'' in \emph{IJCAI}, vol.
  2016, 2016, pp. 3889--3895.

\bibitem{qiu2018unifying}
J.~Qiu, Y.~Dong, H.~Ma, J.~Li, K.~Wang, and J.~Tang, ``Network embedding as
  matrix factorization: Unifying deepwalk, line, pte, and node2vec,'' in
  \emph{Proceedings of the Eleventh ACM International Conference on Web Search
  and Data Mining}, 2018, pp. 459--467.

\bibitem{chen2019deepwalk}
F.~Chen, B.~Wang, and C.-C.~J. Kuo, ``Deepwalk-assisted graph pca (dgpca) for
  language networks,'' in \emph{ICASSP 2019-2019 IEEE International Conference
  on Acoustics, Speech and Signal Processing (ICASSP)}.\hskip 1em plus 0.5em
  minus 0.4em\relax IEEE, 2019, pp. 2957--2961.

\bibitem{chen2020prediction}
Z.-H. Chen, Z.-H. You, Z.-H. Guo, H.-C. Yi, G.-X. Luo, and Y.-B. Wang,
  ``Prediction of drug--target interactions from multi-molecular network based
  on deep walk embedding model,'' \emph{Frontiers in Bioengineering and
  Biotechnology}, vol.~8, p. 338, 2020.

\bibitem{chen2018harp}
H.~Chen, B.~Perozzi, Y.~Hu, and S.~Skiena, ``Harp: Hierarchical representation
  learning for networks,'' in \emph{Proceedings of the AAAI Conference on
  Artificial Intelligence}, vol.~32, no.~1, 2018.

\bibitem{pimentel2017unsupervised}
T.~Pimentel, A.~Veloso, and N.~Ziviani, ``Unsupervised and scalable algorithm
  for learning node representations,'' \emph{International Conference of
  Learning Representation}, 2017.

\bibitem{yang2015multi}
Z.~Yang, J.~Tang, and W.~Cohen, ``Multi-modal bayesian embeddings for learning
  social knowledge graphs,'' \emph{International Joint Conference on Artificial
  Intelligence}, 2015.

\bibitem{pan2016tri}
S.~Pan, J.~Wu, X.~Zhu, C.~Zhang, and Y.~Wang, ``Tri-party deep network
  representation,'' \emph{International Joint Conference on Artificial
  Intelligence}, 2016.

\bibitem{li2016discriminative}
J.~Li, J.~Zhu, and B.~Zhang, ``Discriminative deep random walk for network
  classification,'' in \emph{Proceedings of the 54th Annual Meeting of the
  Association for Computational Linguistics (Volume 1: Long Papers)}, 2016, pp.
  1004--1013.

\bibitem{tang2015line}
J.~Tang, M.~Qu, M.~Wang, M.~Zhang, J.~Yan, and Q.~Mei, ``Line: Large-scale
  information network embedding,'' in \emph{Proceedings of the 24th
  international conference on world wide web}, 2015, pp. 1067--1077.

\bibitem{cao2015grarep}
S.~Cao, W.~Lu, and Q.~Xu, ``Grarep: Learning graph representations with global
  structural information,'' in \emph{Proceedings of the 24th ACM international
  on conference on information and knowledge management}, 2015, pp. 891--900.

\bibitem{chen2019fastrp}
H.~Chen, S.~F. Sultan, Y.~Tian, M.~Chen, and S.~Skiena, ``Fast and accurate
  network embeddings via very sparse random projection,'' in \emph{Proceedings
  of the 28th ACM International Conference on Information and Knowledge
  Management}, ser. CIKM '19.\hskip 1em plus 0.5em minus 0.4em\relax New York,
  NY, USA: Association for Computing Machinery, 2019, p. 399–408.

\bibitem{qiu2020matrix}
J.~Qiu, C.~Wang, B.~Liao, R.~Peng, and J.~Tang, ``A matrix chernoff bound for
  markov chains and its application to co-occurrence matrices,'' \emph{Advances
  in Neural Information Processing Systems}, vol.~33, 2020.

\bibitem{zhang2021consistency}
Y.~Zhang and M.~Tang, ``Consistency of random-walk based network embedding
  algorithms,'' \emph{arXiv preprint arXiv:2101.07354}, 2021.

\bibitem{mikolov2013negativesampling}
T.~Mikolov, I.~Sutskever, K.~Chen, G.~S. Corrado, and J.~Dean, ``Distributed
  representations of words and phrases and their compositionality,'' in
  \emph{Advances in neural information processing systems}, 2013, pp.
  3111--3119.

\bibitem{norris1997markovchainsbook}
J.~R. Norris, \emph{Markov Chains}, ser. Cambridge Series in Statistical and
  Probabilistic Mathematics.\hskip 1em plus 0.5em minus 0.4em\relax Cambridge
  University Press, 1997, p. 1–59.

\bibitem{lovasz1993randomwalkssurvey}
L.~Lov{\'a}sz, ``Random walks on graphs: A survey,'' \emph{Combinatorics, Paul
  Erd\H{o}s is Eighty}, vol.~2, no.~1, pp. 1--46, 1993.

\bibitem{levin2017markovmixing}
D.~A. Levin, Y.~Peres, and E.~L. Wilmer, \emph{Markov Chains and Mixing
  Times}.\hskip 1em plus 0.5em minus 0.4em\relax American Mathematical Soc.,
  2017, vol. 107.

\bibitem{diaconis1991geometric}
P.~Diaconis and D.~Stroock, ``Geometric bounds for eigenvalues of markov
  chains,'' \emph{The Annals of Applied Probability}, pp. 36--61, 1991.

\bibitem{tang2009relational}
L.~Tang and H.~Liu, ``Relational learning via latent social dimensions,'' in
  \emph{Proceedings of the 15th ACM SIGKDD international conference on
  Knowledge discovery and data mining}, 2009, pp. 817--826.

\bibitem{sen2008collective}
P.~Sen, G.~Namata, M.~Bilgic, L.~Getoor, B.~Galligher, and T.~Eliassi-Rad,
  ``Collective classification in network data,'' \emph{AI magazine}, vol.~29,
  no.~3, pp. 93--93, 2008.

\bibitem{rozemberczki2019multi}
B.~Rozemberczki, C.~Allen, and R.~Sarkar, ``Multi-scale attributed node
  embedding,'' \emph{arXiv preprint arXiv:1909.13021}, 2019.

\bibitem{mislove2007measurement}
A.~Mislove, M.~Marcon, K.~P. Gummadi, P.~Druschel, and B.~Bhattacharjee,
  ``Measurement and analysis of online social networks,'' in \emph{Proceedings
  of the 7th ACM SIGCOMM conference on Internet measurement}, 2007, pp. 29--42.

\end{thebibliography}
\bibliographystyle{IEEEtran}





\end{document}